\documentclass{article}

\usepackage{microtype}
\usepackage{graphicx}
\usepackage{subcaption}
\usepackage{booktabs} 

\usepackage{hyperref}

\usepackage[preprint]{icml2026}

\usepackage{amsmath}
\usepackage{amssymb}
\usepackage{mathtools}
\usepackage{amsthm}
\usepackage{caption}

\usepackage[capitalize,noabbrev]{cleveref}

\theoremstyle{plain}

\theoremstyle{definition}

\theoremstyle{remark}

\usepackage[textsize=tiny]{todonotes}

\icmltitlerunning{MLLM-4D: Towards Visual-based Spatial-Temporal Intelligence}

\def\ie{\emph{i.e.}}
\def\eg{\emph{e.g.}}
\def\etc{\emph{etc.}} 
\def\cf{\emph{cf.}} 
\def\vs{\emph{vs.}} 
\def\wrt{\emph{w.r.t.}} 
\def\etc{\emph{etc.}} 
\def\etal{\emph{et al.}} 

\usepackage{enumitem}
\usepackage{amsfonts}       
\usepackage{multirow}
\usepackage{bm}
\usepackage{float}
\usepackage{url} 
\usepackage{mathrsfs}
\usepackage{caption}
\usepackage{xspace}
\usepackage{xcolor}
\usepackage[most]{tcolorbox}
\usepackage[table,dvipsnames]{xcolor}

\begin{document}

\twocolumn[
  \icmltitle{MLLM-4D: Towards Visual-based Spatial-Temporal Intelligence}



  \icmlsetsymbol{equal}{*}
  \icmlsetsymbol{correspondingauthor}{$\dagger$}

  \begin{icmlauthorlist}
    \icmlauthor{Xingyilang Yin}{sch1,sch4,equal}
    \icmlauthor{Chengzhengxu Li}{sch2,equal}
    \icmlauthor{Jiahao Chang}{sch3}
    \icmlauthor{Chi-Man Pun}{sch1,correspondingauthor}
    \icmlauthor{Xiaodong Cun}{sch4,correspondingauthor}
  \end{icmlauthorlist}

  \icmlaffiliation{sch1}{University of Macau}
  \icmlaffiliation{sch2}{Xi'an Jiaotong University}
  \icmlaffiliation{sch3}{The Chinese University of Hong Kong, Shenzhen}
  \icmlaffiliation{sch4}{GVC Lab, Great Bay University}
  \icmlcorrespondingauthor{Chi-Man Pun}{cmpun@um.edu.mo}
  \icmlcorrespondingauthor{Xiaodong Cun}{cun@gbu.edu.cn}

  \icmlkeywords{Machine Learning, ICML}
  
  \vskip 0.1in
  {
    \includegraphics[width=\textwidth]{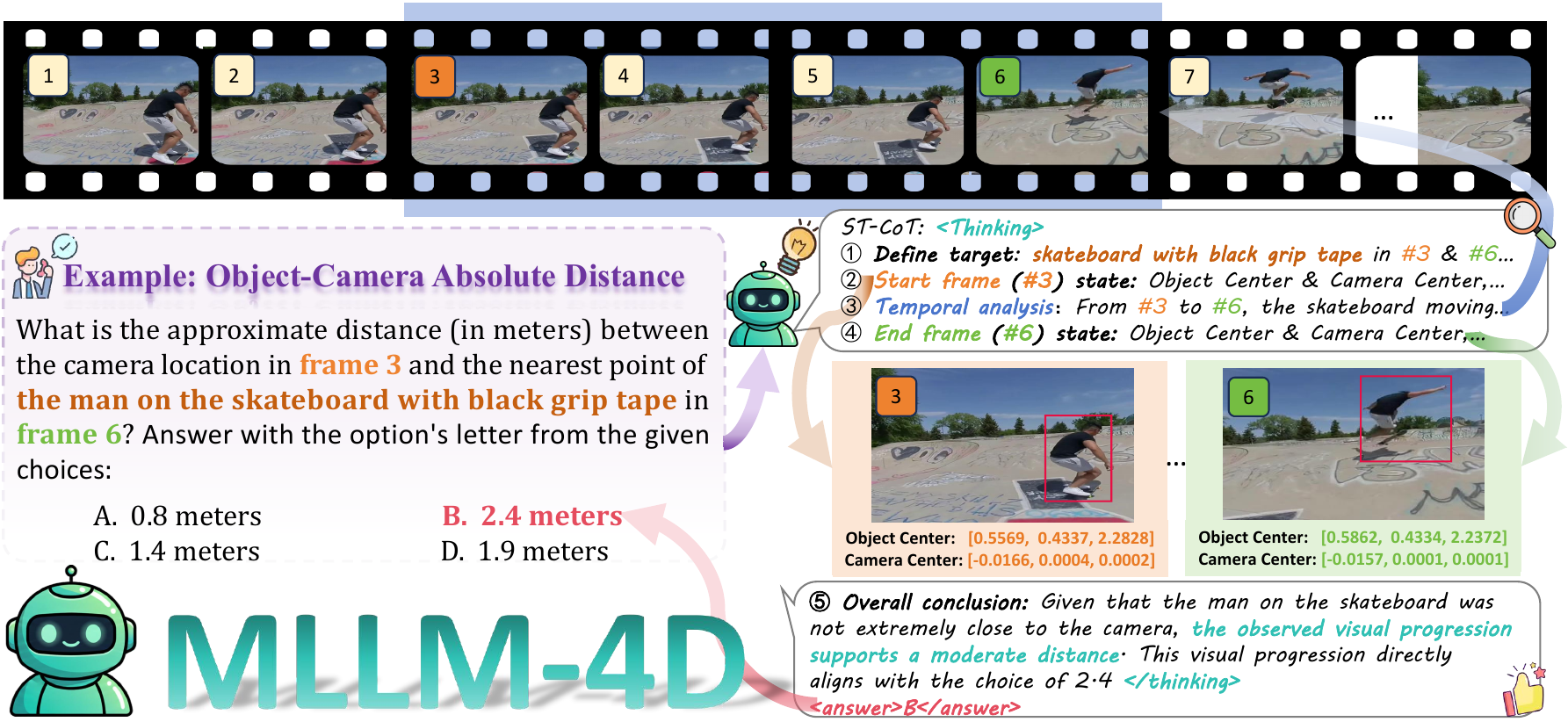}
    \vspace{-0.3em}
    \captionsetup[figure]{hypcap=false}
    \captionof{figure}{
    We propose \textit{MLLM-4D}, a method that advances MLLMs for the visual-based spatial-temporal intelligence. MLLM-4D is capable of understanding and reasoning about the evolution of 3D space over time from only 2D video input.
    }
    \vspace{-0.5em}
    \label{fig:teaser}%
    }
  \vskip 0.3in
    
]



\printAffiliationsAndNotice{\icmlEqualContribution}

\newcommand{\xiaodong}[1]{{\textcolor{orange}{[xd: #1]}}}
\newcommand{\xd}[1]{{\textit{\textcolor{orange}{{#1}}}}}

\makeatletter
\DeclareRobustCommand\onedot{\futurelet\@let@token\@onedot}
\def\@onedot{\ifx\@let@token.\else.\null\fi\xspace}

\def\eg{\emph{e.g}\onedot} \def\Eg{\emph{E.g}\onedot}
\def\ie{\emph{i.e}\onedot} \def\Ie{\emph{I.e}\onedot}
\def\cf{\emph{c.f}\onedot} \def\Cf{\emph{C.f}\onedot}
\def\etc{\emph{etc}\onedot} \def\vs{\emph{vs}\onedot}
\def\wrt{w.r.t\onedot} \def\dof{d.o.f\onedot}
\def\etal{\emph{et al}\onedot}
\makeatother

\begin{abstract}
Humans are born with vision-based 4D spatial-temporal intelligence, which enables us to perceive and reason about the evolution of 3D space over time from purely visual inputs. Despite its importance, this capability remains a significant bottleneck for current multimodal large language models (MLLMs). To tackle this challenge, we introduce \textbf{\textit{MLLM-4D}}, a comprehensive framework designed to bridge the gaps in \textbf{\textit{training data curation}} and \textbf{\textit{model post-training}} for spatiotemporal \textit{understanding} and \textit{reasoning}. On the data front, we develop a cost-efficient data curation pipeline that repurposes existing stereo video datasets into high-quality 4D spatiotemporal instructional data. This results in the \textit{MLLM4D-2M} and \textit{MLLM4D-R1-30k} datasets for Supervised Fine-Tuning (SFT) and Reinforcement Fine-Tuning (RFT), alongside \textit{MLLM4D-Bench} for comprehensive evaluation. Regarding model training, our post-training strategy establishes a foundational 4D understanding via SFT and further catalyzes 4D reasoning capabilities by employing Group Relative Policy Optimization (GRPO) with specialized \textit{Spatiotemporal Chain of Thought (ST-CoT)} prompting and \textit{Spatiotemporal reward functions (ST-reward)} without involving the modification of architecture. Extensive experiments demonstrate that MLLM-4D achieves state-of-the-art spatial-temporal understanding and reasoning capabilities from purely 2D RGB inputs. Project page: \url{https://github.com/GVCLab/MLLM-4D}.
\end{abstract}  
\vspace{-2em}
\section{Introduction}
\vspace{-0.5em}
Humans possess an innate 4D spatiotemporal intelligence that extends beyond the perception of static 3D geometry to integrate time as an intrinsic cognitive dimension. This ability to reason about the world in 4D (3D space + time) allows us to navigate and act effectively within constantly changing environments using purely visual inputs. Such ability is critical to interactive AI systems, including robotics, VR, and embodied agents, where navigating dynamic scenes requires a continuous understanding of evolving spatial relationships. While Multimodal Large Language Models~(MLLMs) have demonstrated remarkable general intelligence~\cite{hurst2024gpt, comanici2025gemini} in image~\cite{liu2023visual, li2024llava}, video~\cite{Qwen2.5-VL, Qwen3-VL}, and audio~\cite{chu2024qwen2, xu2025qwen2}, their abilities for this spatiotemporal understanding and reasoning remain largely underexplored.

Recent works mainly focus on spatial reasoning within static scenes~\cite{yang2025thinking, dihan2025mapeval, yang2025mmsi, azuma2022scanqa, ma2022sqa3d, zhang2025from} and struggle to understand and reason about evolving relationships within 4D space, as illustrated in Fig.~\ref{fig:teaser}. For dynamic scenarios, manual annotations can only collect small benchmark-size datasets~\cite{zhou2025vlm4d,li2025sti} and are challenging to scale for current MLLM training. On the other hand, current methods enhance 3D spatial intelligence of MLLMs using additional spatial encoders~\cite{zheng2025learning, fan2025vlm, huang2024chat, deng20253d, zhu2025llava, liu2025ssr}. However, these 3D expertise MLLMs often fail in dynamic reasoning tasks, as their learned knowledge is constrained to static environments with immobile objects.

We thus introduce \textit{MLLM-4D}, a novel and comprehensive framework for boosting MLLM capabilities for visual-based spatiotemporal intelligence by addressing two critical bottlenecks introduced above: \textit{(i)~ the scarcity of large-scale and high-quality 4D instructional data.} We develop an automated data engine that repurposes existing stereo video datasets~\cite{shao2024deepseekmath} into high-quality 4D spatiotemporal instructional data. Our pipeline integrates several advanced vision primitives to decompose these scenes into per-frame camera poses, object-level 3D points and corresponding semantic descriptions, capturing the rich 4D evolution of the entire scene. By applying rigorous physics-based spatiotemporal computations to this metadata, we generate a large-scale \textit{MLLM4D-2M} dataset for Supervised Fine-Tuning (SFT), \textit{MLLM4D-R1-30k} dataset for Reinforcement Fine-Tuning (RFT), and \textit{MLLM4D-Bench} for comprehensive evaluation, respectively. \textit{(ii)~ the lack of specialized and scalable 4D-aware MLLMs.} Contrary to prior works that rely on auxiliary spatial encoders, we demonstrate that standard MLLM architectures can achieve robust spatiotemporal reasoning when scaled with high-quality 4D data. Our framework utilizes a hierarchical post-training approach to associate pixel-level video observations with 4D physical reasoning. In the first stage, SFT establishes a foundational 4D understanding, ensuring the model can correctly identify spatial-temporal anchors. In the second stage, we catalyze advanced 4D reasoning capabilities through Group Relative Policy Optimization (GRPO)~\cite{liu2024deepseek}. We introduce a specialized five-step \textit{Spatiotemporal Chain of Thought (ST-CoT)} prompting that forces the model to act as a visual physics engine, focusing on temporal anchoring, 3D state parsing, and physical motion. We move beyond standard accuracy reward and format reward by introducing \textit{Spatiotemporal reward (ST-reward)} functions. This reward serves as a physical regularizer, penalizing the model for hallucinated motion that contradicts the actual spatiotemporal evolution of the scene. Extensive experiments demonstrate that MLLM-4D achieves state-of-the-art 4D spatiotemporal understanding and reasoning performance. 

Our main contributions are summarized as follows:
\begin{itemize}[leftmargin=*]
    \item We introduce \textit{MLLM-4D}, a novel and comprehensive framework that significantly enhances the spatial-temporal intelligence of MLLMs, demonstrating strong 4D understanding and reasoning capabilities without requiring architectural modifications.
    
    \item We develop an \textit{automated data curation pipeline} to generate high-quality 4D spatiotemporal instructional data by repurposing the existing stereoscopic video datasets. Leveraging this pipeline, we propose the \textit{MLLM4D-2M} and \textit{MLLM4D-R1-30k} datasets for SFT and RFT, alongside \textit{MLLM4D-Bench} for comprehensive evaluation.

    \item In our training framework, we propose specialized \textit{ST-CoT} prompting strategies and physics-grounded \textit{ST-reward}. These are integrated into GRPO to systematically improve the model's capacity for verifiable 4D spatiotemporal reasoning in dynamic scenes.

    \item Extensive experiments demonstrate that our MLLM-4D achieves state-of-the-art 4D understanding and reasoning performance with only RGB video input.

\end{itemize}
 
\vspace{-1em}
\section{Related Works}
\vspace{-0.5em}
\noindent\textbf{Multimodal Large Language Models .}
Multimodal Large Language Models (MLLMs)~\cite{liu2023visual, li2023blip, zhang2024llavanext-video, hurst2024gpt, tong2024cambrian, wang2025internvideo2, zhou2025strefer, comanici2025gemini} have achieved remarkable success across diverse 2D visual tasks. Recent advancements, such as Qwen3-VL~\cite{Qwen3-VL} achieve strong visual modeling across images and video by integrating interleaved-MRoPE, multi-level visual features, and text-based time alignment. Despite these strides, even the state-of-the-art MLLMs struggle to interpret the complex underlying 4D scene from video. Our proposed MLLM-4D is designed to bridge this gap, achieving visual-based spatiotemporal intelligence inspired by human innate cognition. Much like the innate ability of human brain to perceive 2D visual signals yet reason about 3D physical relationships over time, MLLM-4D enables deeper structural understanding of the dynamic world.

\noindent\textbf{MLLMs for Spatial Intelligence.}
Recent advances~\cite{ma2025spatialreasoner, shen2025fine, xu2025multi, ouyang2025spacer, yangtowards} have sparked interest in extending MLLMs to encompass 3D spatial understanding and reasoning. While some methods rely on auxiliary 3D geometric input~\cite{huang2024chat, deng20253d} or 2.5D depth information~\cite{zhu2025llava, liu2025ssr}, recent studies including VG-LLM~\cite{zheng2025learning}, Spatial-MLLM~\cite{wu2025spatial}, and VLM-3R~\cite{fan2025vlm} attempt to perceive the 3D world directly from video by leveraging 3D reconstruction priors~\cite{wang2025vggt, wang2025continuous}. However, these 3D expertise MLLMs remain largely constrained to static scenes with immobile objects and struggle to learn the evolving relationships within a 4D spatiotemporal manifold. In contrast, our MLLM-4D establishes a foundational 4D understanding and reasoning capabilities with our dataset and designed training recipes.

\vspace{-0.4em}
\noindent\textbf{Visual-based 4D Spatial-Temporal Intelligence.}
Visual-based spatial-temporal intelligence focuses on enabling video MLLMs to understand and reason about 4D spatiotemporal relationships directly from visual input. Previous works mainly focus on improving the spatial intelligence of MLLMs through 3D QA datasets~\cite{azuma2022scanqa, ma2022sqa3d, zhang2025from} on static 3D spatial reasoning benchmarks~\cite{yang2025thinking, dihan2025mapeval, yang2025mmsi, jia2025omnispatial}. More recently, efforts~\cite{zhou2025vlm4d, li2025sti} such as VLM4D~\cite{zhou2025vlm4d} have begun to evaluate the spatiotemporal reasoning capabilities of MLLMs. However, these benchmarks are limited to a few thousand QA pairs and rely on manual annotation, which lacks the scalability required for MLLM fine-tuning. To address the data constraints, we propose an automated data curation pipeline to generate large-scale training and evaluation datasets, establishing a high-quality foundation for 4D spatiotemporal intelligence learning.

\vspace{-0.4cm}
\section{Scalable Spatial-Temporal Data Curation}\label{sec:scalable st data curation}
\vspace{-0.2cm}
We utilize the question-answer~(QA) pair as the fundamental training type to enhance the spatial-temporal understanding and reasoning of MLLM. However, current 4D instructional datasets~\cite{zhou2025vlm4d, li2025sti} are typically limited to benchmark-scale samples (e.g, about 2k in VLM4D), as they rely on manual annotations with unstructured types, which are not scalable for training MLLMs.

Thus, we first define the included scope of the spatial-temporal intelligence from the decoupling aspect of camera and object movement. As shown in Fig.~\ref{fig: MLLM-Bench}, which includes three distinct categories: 

\begin{figure}[t]
  \centering
   \includegraphics[width=\linewidth]{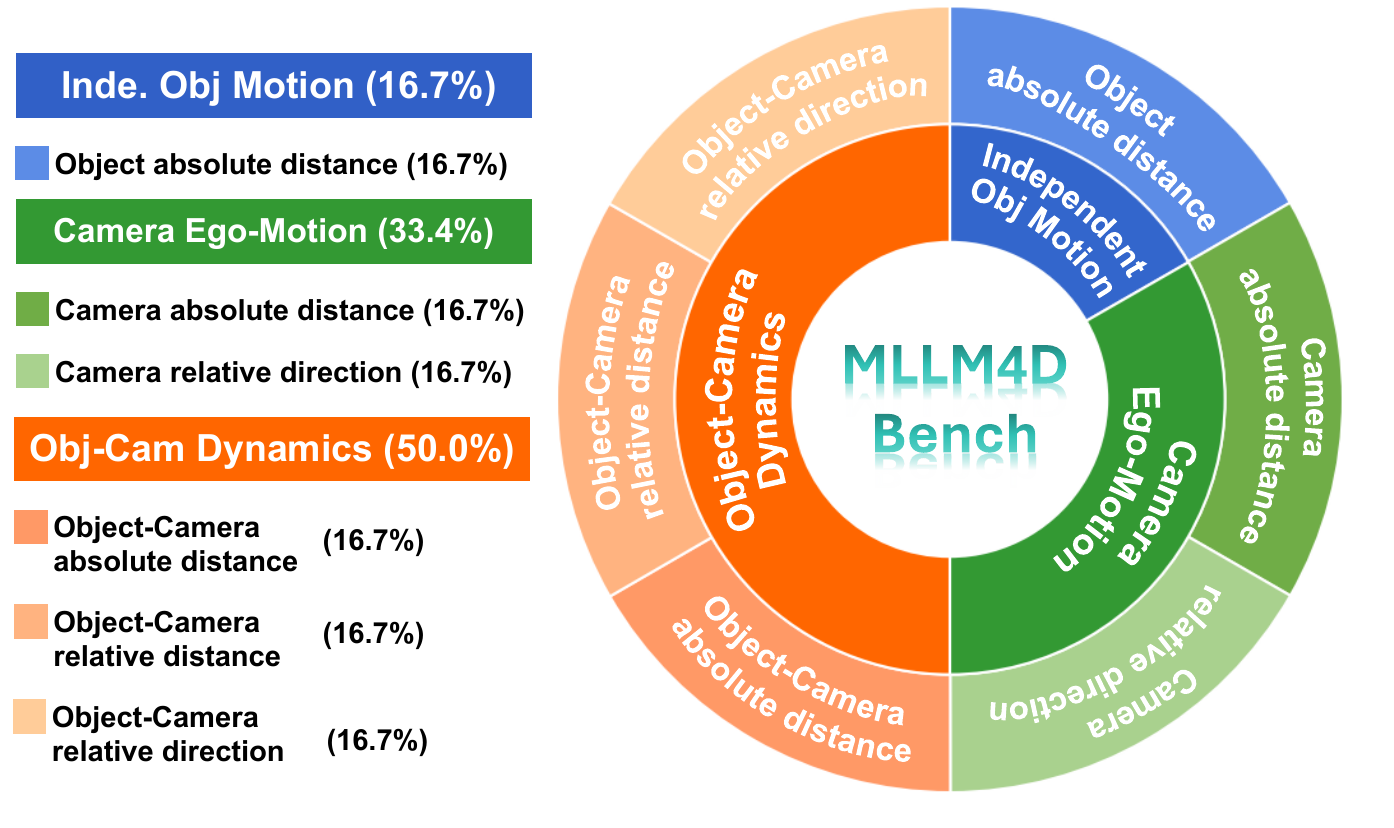}
   \vspace{-1.8em}
   \caption{The components of our MLLM4D-Bench.}
   \vspace{-2.2em}
   \label{fig: MLLM-Bench}
\end{figure}

\textit{(i) Independent Object Motion} has one type of question, which aims to ask about the objects' absolute distance changes over time (frames).

\textit{(ii) Camera Ego-Motion} assesses the camera's displacement and orientation changes through two metrics: \textit{camera absolute distance}, which measures the physical span of the camera's movement among frames, and \textit{camera relative direction}, which tracks the angular trajectory of the camera over temporal dimension (frames).

\textit{(iii) Object-Camera Dynamics.} This category characterizes the intricate spatial interplay between the camera and moving objects. It encompasses three core dimensions: the absolute distance between the two, their relative distance change, and their relative angular orientation.

To automatically obtain the scalable and accurate question-answer pairs, we predict the spatial-temporal metadata from the video and obtain the answer via physical laws. 
This metadata contains the per-frame camera poses, object-level 3D point clouds, and fine-grained semantic descriptions, which constitutes a comprehensive 4D representation of dynamic scenes. However, directly processing monocular videos through 4D tracking models~\cite{xiao2025spatialtrackerv2, badki2025l4p} often suffers from depth ambiguity and accuracy issues. Instead, we introduce an automated pipeline that extracts precise 4D spatiotemporal information from existing stereoscopic video datasets~\cite{jin2025stereo4d}. After obtaining the metadata, we utilize the physical-based formulations to solve for exact spatiotemporal relationships and obtain the thinking process via the MLLM-based CoT generation pipeline, alongside the MLLM4D-Bench for evaluation. Below, we give the details of each part.

\begin{figure*}[t]
  \centering
   \includegraphics[width=\linewidth]{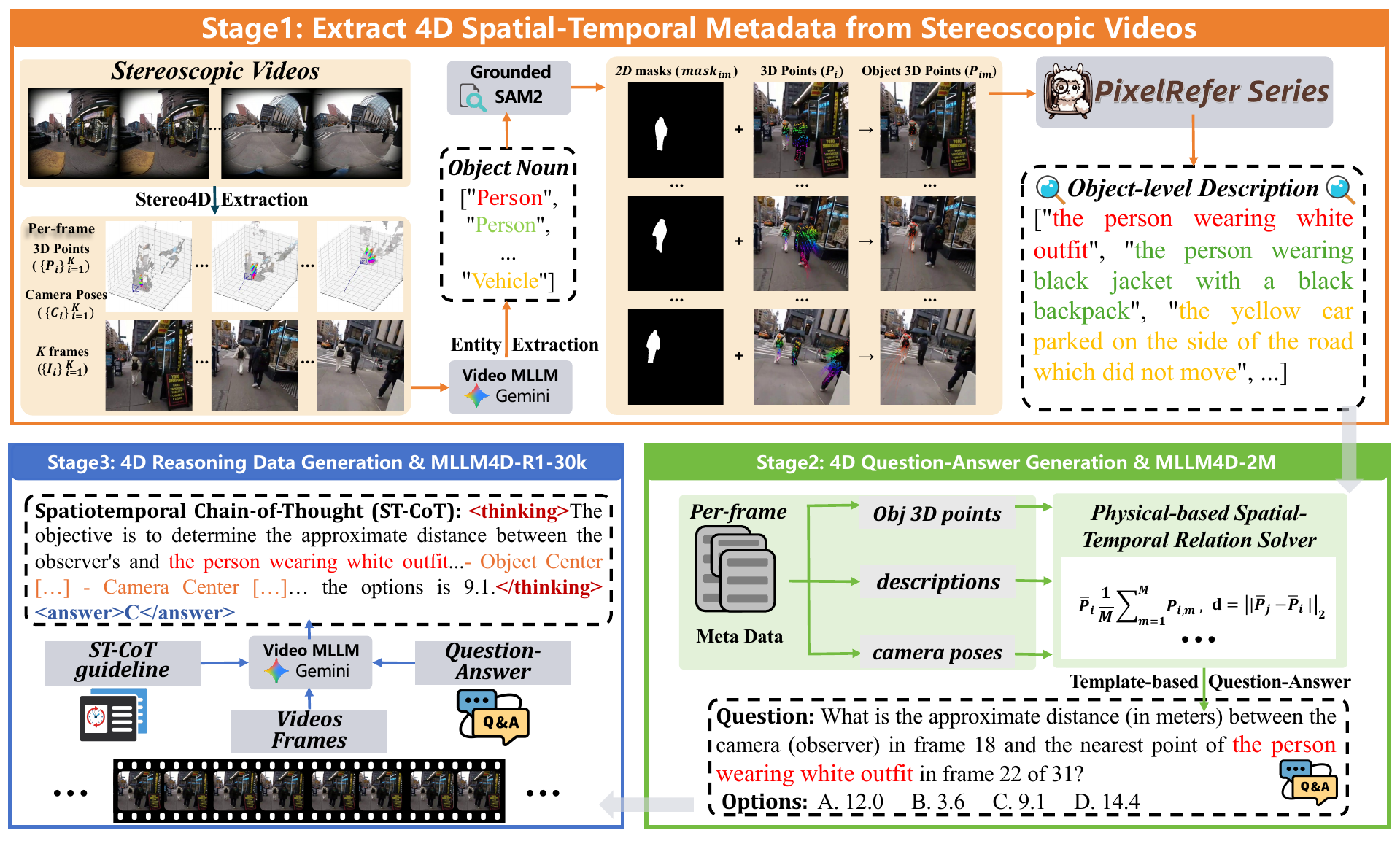}
   \vspace{-1.8em}
   \caption{Our scalable curation pipeline for instructional spatiotemporal data. Our automated pipeline leverages several advanced vision techniques to extract 4D spatiotemporal information from stereoscopic videos, including per-frame camera poses, object-level 3D point clouds, and semantic descriptions. These data are then processed through a physics-based spatiotemporal relation solver to generate 4D QA pairs, and our specialized ST-CoT prompting strategy synthesizes the corresponding reasoning trajectories.}
   \vspace{-1.1em}
   \label{fig2: 4d spatial-temporal data pipeline}
\end{figure*}

\noindent\textbf{4D Spatial-Temporal Metadata from Stereo Videos.}
As illustrated in Fig.~\ref{fig2: 4d spatial-temporal data pipeline}, giving the stereoscopic video from Stereo4D~\cite{jin2025stereo4d}, we obtain the $K$ left-rectified video frames $\{I_i\}_{i=1}^{K}$; processed camera pose~\cite{schonberger2016structure} $\{C_i\}_{i=1}^{K}$, where $C_i = [R_i | t_i]$ consists of a $3\times3$ rotation matrix $R_i$ and a $3\times1$ translation vector $t_i$; per-frame metric 3D points~\cite{doersch2024bootstap} $\{P_i\}_{i=1}^{K}$ from metric stereo depth~\cite{wang2024sea} to obtain the stereo metadata. This is followed by a robust filtering stage to filter out the low-quality estimations.

After that, to obtain the instance-level semantic annotation from a question-answer pair, we employ a video MLLM, \emph{i.e.} Gemini-2.5-flash~\cite{comanici2025gemini}, to identify all moving entities and extract their corresponding noun categories~(see Appendix~\ref{appendix: Details of MLLM4D-2M} for details). We then utilize GroundedSAM2~\cite{ren2024grounded, ravisam, liu2024grounding} for instance segmentation and tracking, yielding temporally consistent 2D masks across the sequence. Finally, the scene-level 4D points $\{P_i\}_{i=1}^{K}$ are projected onto these 2D masks $\{mask_{im}\}_{i=1,...,K; m=1,...,M}$ to isolate per-frame, object-level 4D points for M distinct objects, denoted as $\{P_{im}\}_{i=1,...,K; m=1,...,M}$. To enrich these representations, the video frames and their associated 2D masks are fed into a region-level MLLM, PixelRefer~\cite{yuan2025pixelrefer}, to generate fine-grained semantic descriptions $\{T_m\}_{m=1}^{M}$ for each object. The camera pose, fine-grained description, and the instance-level 3D points build the general form of the metadata for further spatial-temporal relationship solver via physical laws.

\begin{figure*}[t]
  \centering
   \includegraphics[width=\linewidth]{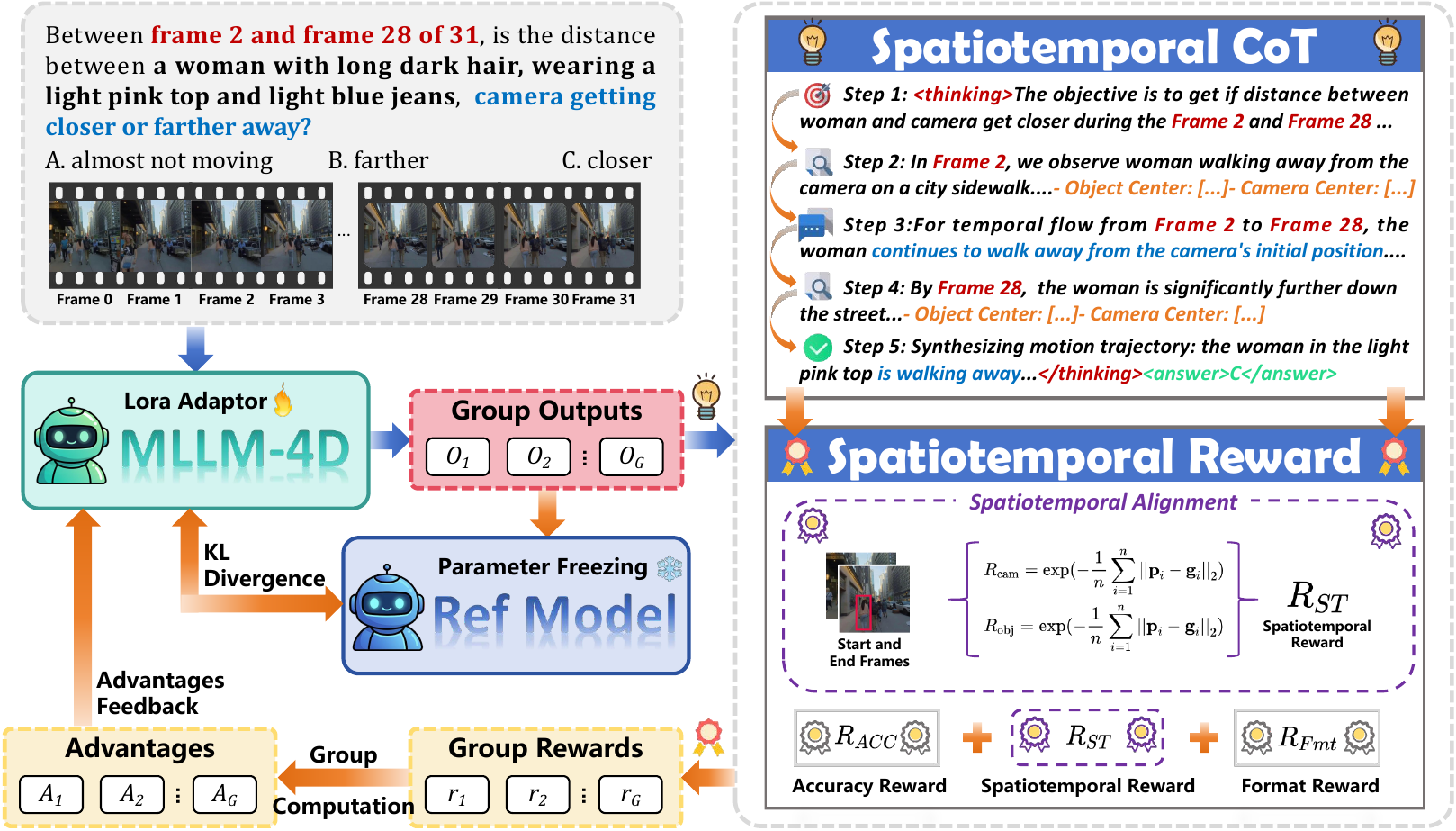}
   \vspace{-1.3em}
   \caption{Our RFT pipeline. Given the input video and question, the MLLM-4D model generates multiple rollouts using the ST-CoT reasoning format. Within each group, relative advantages are computed based on accuracy reward, format reward and ST-reward. The model parameters are then updated via the GRPO objective, which incorporates a KL penalty relative to the frozen reference model.}
   \vspace{-1.1em}
   \label{fig3: GRPO}
\end{figure*}

\noindent\textbf{Physical-based Spatial-Temporal Relationship Solver.} To generate high-fidelity QA pairs, we derive ground-truth results through rigorous physics-based spatial-temporal computation. For instance, to determine the Camera Relative Direction between any two frames $i$ and $j$: we first compute the world-space displacement $\Delta t = t_j - t_i$. To determine the movement relative to the camera's perspective at frame $i$, we project this vector into the local camera coordinate system: $d_{cam\_rel\_dir} = R_i^T \cdot \Delta t$. The resulting vector $d_{cam\_rel\_dir} = [dx, dy, dz]$ is interpreted according to standard camera conventions, where $+X, +Y, +Z$ correspond to the right, down, and forward directions, respectively. Please refer to Appendix~\ref{appendix: Details of MLLM4D-2M} for more details about other spatiotemporal relation solver implementations.

\noindent\textbf{MLLM4D-2M Dataset.} After calculating the ground-truth values, we apply templates to formulate the final QA pairs. Please refer to the Appendix~\ref{appendix: Details of MLLM4D-2M} for more details about templates and data filtering details. After filtering, we retain $2M$ high-quality QA pairs across approximately $100k$ videos, forming the large-scale \textit{MLLM4D-2M} dataset for supervised fine-tuning (SFT).

\noindent\textbf{MLLM4D-Bench.} We report the composition and evaluation settings of our MLLM4D-Bench in Fig.~\ref{fig: MLLM-Bench}, which comprise 6k questions organized into six specialized subtasks. Our benchmark distinguishes itself from existing evaluation suites in several key dimensions: (1) Dynamic Scene Complexity: Unlike current 3D spatial benchmarks~\cite{yang2025thinking, yang2025mmsi, jia2025omnispatial} which focus on reasoning within static environments, MLLM4D-Bench evaluates dynamic scenes involving both moving camera perspectives and multiple moving objects; (2) Structured 4D Motion Categorization: Compared to existing 4D benchmarks~\cite{zhou2025vlm4d, li2025sti}, we provide a more rigorous decomposition of 4D motion into three logical categories: Independent Object Motion, Camera Ego-Motion, and Object-Camera Dynamics; (3) Fine-Grained Temporal Evaluation: Our benchmark features fine-grained, frame-wise temporal tagging. This enables the precise evaluation of 4D spatiotemporal reasoning between any two arbitrary frames in a video, which is absent in previous benchmark.

\noindent\textbf{4D Reasoning Data Generation.} 
We synthesize detailed thinking processes to align the model for 4D spatiotemporal reasoning. By prompting Gemini-2.5-Pro with video frames, corresponding QA pairs, 4D physical values, and spatiotemporal CoT guideline, we generate specialized spatiotemporal CoT data (See Sec.~\ref{sec:cot}, Fig.~\ref{fig: ST-CoT prompt} and Appendix~\ref{sec: cold_start details} for details). This process yields 7k cold-start samples and the \textit{MLLM4D-R1-30k}, which contains 30k QA pairs with significant 4D motion and ground-truth solutions designed for large-scale Reinforcement Fine-Tuning.

\vspace{-0.3cm}
\section{MLLM-4D Framework}
\vspace{-0.15cm}
Previous works involve additional spatial encoders~\cite{zheng2025learning, fan2025vlm, wu2025spatial} to boost spatial understanding. Notably, we find that using a \textit{pure 2D visual encoder and retaining the standard MLLM architecture already achieves} state-of-the-art performance when supported by our scalable, high-quality spatial-temporal datasets and optimized post-training framework. We propose a two-stage post-training framework for both understanding and reasoning. In the first stage, we conduct SFT on our MLLM4D-2M dataset to establish foundational 4D understanding~(Sec. \ref{sec:sft}). To further enhance 4D reasoning, we utilize a cold start phase to align the model's output with our specialized \textit{Spatiotemporal Chain of Thought (ST-CoT)}~(Sec.~\ref{sec:cot}). This is followed by the second stage in which we employ Group Relative Policy Optimization (GRPO), leveraging the ST-CoT prompting and our \textit{spatiotemporal reward (ST-reward)} functions on our MLLM4D-R1-30k dataset~(Sec.~\ref{sec:rft}).

\vspace{-1em}
\subsection{Supervised Fine-Tuning for 4D Understanding}\label{sec:sft}
\vspace{-0.5em}
Leveraging the proposed MLLM4D-2M dataset, we first perform supervised fine-tuning~(SFT) to establish foundational 4D spatiotemporal comprehension. To ensure efficient adaptation while preserving the model's pre-trained multimodal knowledge, we utilize the Low-Rank Adaptation (LoRA)~\cite{hulora} technique, where trainable rank-decomposition matrices are injected into the linear layers of Transformers. During this stage, we employ the standard cross-entropy loss $\mathcal{L}_{\text{ce}}$ between the model-generated answer sequences and the ground-truth annotations:
\vspace{-0.8em}
\begin{equation}\label{eq:loss_ce}
\mathcal{L}_{\text{ce}} = - \sum_{j} \log P(o^{(j)} \mid o^{(1:j-1)}, q, \{I_i\}_{i=1}^{N_k}),
\vspace{-0.8em}
\end{equation}
where $\{I_i\}_{i=1}^{N_k}$ denotes the sequence of $N_k$ input video frames, $q$ represents the concatenated system prompt and scenario-specific question, $o^{(i)}$ denotes the $i$-th token in the target reasoning-answer trajectory, and $o^{(1:i-1)}$ denotes the preceding tokens. This foundational phase ensures the model internalizes the prerequisite spatial-temporal alignment necessary for subsequent high-level reasoning.

\vspace{-1.1em}
\subsection{Cold-Start Alignment for 4D Reasoning}\label{sec:cot}
\vspace{-0.5em}
As shown in Fig.~\ref{fig3: GRPO}, to transition from basic perception to complex 4D analysis, we introduce the \textit{Spatiotemporal Chain-of-Thought (ST-CoT)}, a reasoning paradigm designed to \textbf{associate 2D pixel-level observations with 4D world-state representations} (See Fig.~\ref{fig: ST-CoT prompt} for detail prompting settings). Unlike conventional CoT approaches \cite{wei2022chain} that primarily navigate linguistic logic, ST-CoT compels the model to operate as a 3D perception engine, grounding its internal reasoning in visual physics and motion dynamics over time. We utilize this framework to curate a specialized cold-start dataset (see Appendix~\ref{sec: cold_start details} for details). By performing cold-start alignment on this structured data using the same LoRA parameter and loss objective $\mathcal{L}_{\text{ce}}$ as the SFT stage, we establish a policy initialization that serves as a stable foundation for subsequent reinforcement learning. The ST-CoT guideline follows the five-step logical flow:

\vspace{-0.3em}
\textit{Step 1: Objective Alignment and Temporal Anchoring:} The reasoning process begins by explicitly defining the spatiotemporal objective $O_{obj}$. 
The model identifies the query's intent and locks onto critical temporal boundaries (the \textit{start frame $t_{start}$} and \textit{end frame $t_{end}$}), which prevent computational drift during video processing. 

\vspace{-0.3em}
\textit{Step 2: Start frame 3D State Parsing and Anchoring:} At the start frame $t_{start}$, the model performs a joint visual and geometric analysis to output the initial spatial state $\mathcal{S}_{t_{start}}$. By anchoring the camera center, object center, and scene descriptions, the model establishes a quantitative baseline for all subsequent motion estimations.

\vspace{-0.3em}
\textit{Step 3: Temporal Progression and Visual Cue Collection:} Rather than relying on implicit visual flow, the model rationalizes the physical shift between boundaries. It analyzes visual transformations, such as scale expansion and perspective distortion, to infer underlying geometric deltas $\mathcal{T}_{\text{motion}}$, creating a causal bridge between visual observation and 4D physical motion.

\vspace{-0.3em}
\textit{Step 4: End frame 3D State Verification:} Upon reaching $t_{end}$, the model generates the terminal spatial state $\mathcal{S}_{t_{end}}$ and a corresponding visual summary. This serves as a consistency check; by comparing $\mathcal{S}_{t_{e}}$ with $\mathcal{S}_{t_{s}}$, the model validates the continuity of the spatiotemporal trajectory and minimizes temporal hallucinations.

\vspace{-0.3em}
\textit{Step 5: Evidence-Based Synthesis and Probabilistic Inference:} Finally, the model synthesizes the accumulated visual evidence into a comprehensive trajectory. We formally define the ST-CoT as the coherent sequence $\mathcal{T}_{ST} = \{O_{obj}, \mathcal{S}_{s}, \mathcal{T}_{motion}, \mathcal{S}_{e} \}$, aggregating the objective $O_{obj}$, spatial anchors ($\mathcal{S}_{s}$ \& $\mathcal{S}_{e}$), and temporal motion ($\mathcal{T}_{motion}$). The final answer $\hat{a}$ is derived through a maximum a posteriori estimation, conditioned strictly on this reconstructed 4D trajectory and the raw video input $\mathcal{V}$:
\vspace{-1em}
\begin{equation} \hat{a} = \arg\max_{a \in \mathcal{A}} P(a \mid \underbrace{O_{obj}, \mathcal{S}_{s}, \mathcal{T}_{motion}, \mathcal{S}_{e}}_{\mathcal{T}_{ST}}, \mathcal{V}; \theta).
\vspace{-0.8em}
\end{equation}
This formulation ensures that the final output is not a hallucinated guess but a logical derivative of the explicit 4D progression. By forcing the model to justify its answer through the established trajectory $\mathcal{T}_{ST}$, we enhance the interpretability and capability of the 4D reasoning task.

\begin{table*}[!t]
\centering
\setlength\tabcolsep{2.5pt}
\resizebox{0.95\textwidth}{!}
{
    \begin{tabular}{l | cccccc | c}
        \toprule
         \multirow{3}{*}{\textbf{Models}}  & \multicolumn{6}{c|}{MLLM4D-Bench}  & \multirow{3}{*}{Avg.} \\
         \cmidrule(lr){2-7} 
         & \multicolumn{2}{c}{\textbf{Camera}} & \textbf{Object} & \multicolumn{3}{c|}{\textbf{Object \& Camera}} & \\
         & Abs. Dis & Rel. Dir. & Abs. Dis. & Abs. Dis. & Rel. Dis. & Rel. Dir. &\\
        \midrule
        \rowcolor[RGB]{235, 245, 252}
        & \multicolumn{6}{c}{\textcolor{black}{\textit{Proprietary Models}}} & \\
        GPT-4o~\cite{hurst2024gpt} & 34.8 & 57.6 & 32.7 & 36.7 & 56.7 & 51.0 & 44.9 \\
        Gemini2.5-Flash~\cite{comanici2025gemini} & 35.2 & 59.4 & 24.1 & 37.8 & 55.3 & 50.6 & 43.8 \\
        Gemini2.5-Pro~\cite{comanici2025gemini} & 37.4 & 57.5 & 31.1 & 38.6 & 60.1 & 55.0 & 46.6 \\
        \midrule
        \rowcolor[RGB]{235, 245, 252}
        & \multicolumn{6}{c}{\textcolor{black}{\textit{Open-Source Models}}} & \\
        Qwen2.5-VL-7B~\cite{Qwen2.5-VL} & 25.0 & 58.0 & 24.6 & 25.0 & 42.5 & 30.6 & 34.3 \\
        Qwen2.5-VL-32B~\cite{Qwen2.5-VL} & 32.2 & 52.0 & 31.0 & 30.0 & 3.2 & 9.1 & 26.2 \\
        Qwen3-VL-8B~\cite{Qwen3-VL} & 36.0 & 60.4 & 33.8 & 27.2 & 26.4 & 28.2 & 35.3 \\
        Qwen3-VL-32B~\cite{Qwen3-VL} & 40.4 & 60.8 & 35.1 & 35.2 & 39.5 & 37.0 & 41.3 \\
        LLaVA-NeXT-Video-7B~\cite{zhang2024llavanext-video} & 22.6 & 26.0 & 24.6 & 24.3 & 33.4 & 32.6 & 27.3 \\
        InternVideo2.5-8B~\cite{wang2025internvideo2} & 9.4 & 48.2 & 9.5 & 9.3 & 22.3 & 18.6 & 19.6 \\
        InternVL2.5-8B~\cite{chen2024expanding} & 27.5 & 43.4 & 5.8 & 7.2 & 21.0 & 16.1 & 20.2 \\
        InternVL2.5-38B~\cite{chen2024expanding} & 7.7 & 16.4 & 9.4 & 14.1 & 24.0 & 17.4 & 14.8 \\
        InternVL3.5-8B~\cite{wang2025internvl3} & 34.5 & 55.9 & 33.7 & 34.3 & 41.2 & 36.0 & 39.3 \\
        InternVL3.5-38B~\cite{wang2025internvl3} & 24.3 & 22.6 & 24.0 & 40.9 & 40.7 & 37.1 & 31.6 \\
        \midrule
        \rowcolor[RGB]{235, 245, 252}
        & \multicolumn{6}{c}{\textcolor{black}{\textit{3D Spatial Reasoning Models}}} &  \\
        VLM-3R (LLa.-Video-7B)~\cite{fan2025vlm} & 29.0 & 56.0 & 24.6 & 29.2 & 20.4 & 24.6 & 30.6 \\
        VG-LLM (Qwen2.5-VL-7B)~\cite{zheng2025learning} & 54.9 & 55.7 & 55.6 & 55.7 & 61.8 & 54.3 & 56.3 \\
        \midrule
        \textbf{Our MLLM-4D (Qwen2.5-VL-7B)} & 73.3 & 68.1 & 75.7 & 73.0 & \textbf{70.9} & 60.4 & 70.2 \\
        \textbf{Our MLLM-4D (Qwen3-VL-8B)} & \textbf{73.4} & \textbf{71.9} &\textbf{76.3} & \textbf{74.3} & 69.2 & \textbf{70.9} & \textbf{72.7} \\
        \bottomrule
    \end{tabular}
}
\caption{Comparison of different models on MLLM4D-Bench. Bold font indicates the best performance.}
\vspace{-0.65cm}
\label{tab:mllm4d_bench}
\end{table*}

\vspace{-1em}
\subsection{Reinforcement Fine-Tuning for 4D Reasoning}\label{sec:rft}
\vspace{-0.5em}
Building upon the policy initialization from the cold-start phase, we employ GRPO~\cite{shao2024deepseekmath, liu2024deepseek} to further enhance the model's reasoning capabilities, leveraging the proposed MLLM4D-R1-30k dataset. Unlike traditional Actor-Critic frameworks, GRPO eliminates the need for a separate value function by utilizing the relative rewards within a sampled group. For each input $q$, we sample a group of $G$ outputs $\{o_1, o_2, \dots, o_G\}$ from the current policy $\pi_\theta$. 
The training objective is to maximize the following surrogate loss:
\vspace{-0.9em}
\begin{equation}
\label{eq:loss_grpo}
\small
\begin{aligned}
\mathcal{L}_{grpo} = \mathbb{E} \bigg[ \frac{1}{G} \sum_{g=1}^{G} \Big( & \min \left( \rho_g A_g, \text{clip}(\rho_g, 1-\epsilon, 1+\epsilon) A_g \right) \\
& - \beta \mathbb{D}_{KL}(\pi_\theta \| \pi_{ref}) \Big) \bigg],
\vspace{-1em}
\end{aligned}
\vspace{-1em}
\end{equation}
where $\rho_g = \frac{\pi_\theta(o_g | q)}{\pi_{old}(o_g | q)}$ is the importance sampling ratio, and $\pi_{ref}$ is the reference model (the cold-start checkpoint in our work). The advantage $A_g$ is computed by normalizing the rewards within the group: $A_g = \frac{r_g - \text{mean}(\mathbf{r})}{\text{std}(\mathbf{r})}$.

\vspace{-0.4em}
While conventional reinforcement learning relies on format and accuracy-based rewards, the specialization of our model for 4D environments hinges upon our proposed Spatiotemporal Reward~(ST-Reward). This reward mechanism serves as a critical supervisor, grounding the ST-CoT reasoning trajectories in precise spatial and temporal physical quantities to ensure logical consistency. As illustrated in Fig.~\ref{fig3: GRPO}, the integration of ST-Reward facilitates the emergence of sophisticated 4D analysis by refining the policy’s ability to generate physically-grounded thoughts.

\vspace{-0.3em}
\noindent\textbf{Accuracy and Format Reward.} The accuracy reward $R_{acc}$ evaluates the correctness of the final prediction, reinforcing alignment with ground-truth labels:
\vspace{-0.9em}
\begin{equation}
    R_{Acc} =  \left\{\begin{matrix} 1 ,& \text{if answer is right}\\ 0 ,& \text{if answer is wrong} \end{matrix}\right..
\vspace{-0.9em}
\end{equation}
The format reward complements the accuracy reward by enforcing strict adherence to the predefined response structure. We utilize regular expressions to verify that the generated trajectories follow the required structure:

\begin{tcolorbox}[
    colback=gray!5, 
    colframe=gray!20, 
    arc=3pt, 
    left=5pt, right=5pt, top=2pt, bottom=2pt
]
\begin{description}[
    leftmargin=2.0em, 
    labelsep=0.3em,   
    nosep,            
    font=\normalfont  
]
    \item[]\texttt{<thinking> Textual Reasoning... Object Center:[...] \\Camera Center:[...] \\Textual Reasoning...</thinking>\\<answer>Final Answer</answer>}.
\end{description}
\end{tcolorbox}

The total format reward is decomposed as:
\vspace{-0.8em}
\begin{equation}
R_{Fmt} = \lambda_{1} R_{Stru\_fmt} + \lambda_{2} R_{ST\_fmt},
\vspace{-0.7em}
\end{equation}
where $R_{Stru\_fmt}\in\{0,1\}$ indicates whether the response is correctly encapsulated within \texttt{<thinking>} and \texttt{<answer>} tags, fostering a standardized CoT workflow. More importantly, we introduce $R_{ST\_fmt}$ rewards the explicit provision of Object Center and Camera Center coordinates in bracketed array formats. This structural constraint serves as a prerequisite, ensuring the reasoning process is parsable for subsequent spatiotemporal reward computation.

\noindent\textbf{Spatiotemporal Reward.}  To constrain reasoning trajectories within a physically plausible 4D world state, we introduce the ST-reward. This reward ensures the model's internal reasoning is grounded in a physically plausible 4D world state rather than mere 2D pixel displacement.  It quantitatively evaluates the model's ability to localize the camera and object at critical temporal anchors, specifically the start and end frames of the video sequence.

The camera and object rewards are computed by mapping the Mean Euclidean Error (MEE) between predicted coordinates $\mathbf{p}_i$ and ground-truth centers $\mathbf{g}_i$ to a normalized range $[0, 1]$ via an exponential decay function:
\vspace{-0.8em}
\begin{equation}
R_{Cam/Obj} = \exp(-\frac{1}{n} \sum_{i=1}^{n} || \mathbf{p}_i - \mathbf{g}_i ||_2)
\vspace{-0.5em}
\end{equation}
where $\| \cdot \|_2$ denotes the $L_2$ norm. A reward of $1.0$ indicates perfect spatial-temporal alignment. Thus the final ST-reward is a composite signal defined as:
\vspace{-0.6em}
\begin{equation}
R_{ST} = \lambda_{Cam} R_{Cam} + \lambda_{Obj} R_{Obj}.
\vspace{-1em}
\end{equation}
By enforcing explicit coordinate prediction, $R_{ST}$ serves as a physical regularizer, effectively mitigating spatiotemporal hallucinations and ensuring the reasoning process adheres to the underlying 4D dynamics. Consequently, the total reward $R$ for GRPO training is defined as:
\vspace{-0.6em}
\begin{equation}
R = \lambda_{Acc} R_{Acc} + \lambda_{Fmt} R_{Fmt} + \lambda_{ST} R_{ST}.
\vspace{-0.8em}
\end{equation} 
\section{Experiments}
\vspace{-0.5em}

\textit{Please refer to the Appendix for the experimental setup, such as implementation details, and the comparison baselines.}

\vspace{-1em}
\subsection{Comparisons on MLLM4D-Bench.}
\vspace{-0.5em}
We compare our MLLM-4D with baselines for 4D spatial-temporal reasoning on MLLM4D-Bench. As shown in Table~\ref{tab:mllm4d_bench}, MLLM-4D significantly outperforms all proprietary models, open-source and 3D spatial reasoning MLLMs. Specifically, our MLLM-4D (Qwen3-VL-8B) variant achieves a state-of-the-art average score of 72.7\%, surpassing high-performing proprietary models like Gemini 2.5 Pro (46.6\%) by a substantial margin. Qualitative comparisons provided in Fig.~\ref{fig: qa_example1}, Fig.~\ref{fig: qa_example2} and Fig.~\ref{fig: qa_example3} further highlight our model's superiority. Beyond simply providing accurate final answers, MLLM-4D demonstrates interpretable 4D reasoning by explicitly modeling the Spatiotemporal Chain-of-Thought (ST-CoT). In contrast, baseline models, including open-source models like Qwen3-VL-8B and 3D spatial reasoning models like VG-LLM, often lack a fundamental understanding of 4D spatiotemporal dynamics. These baselines rely on guesswork or 2D visual cues, leading to incorrect results in 4D reasoning tasks.

\begin{table}[t]
	\centering
    \resizebox{\columnwidth}{!}
    {
	\begin{tabular}{l|ccc}
         \toprule
		 Models & Real & Synthetic & Overall \\
	\hline
    \rowcolor[RGB]{235, 245, 252}
    \multicolumn{4}{c}{\textcolor{black}{\textit{Proprietary Models}}}\\

         GPT-4o & 60.0 & 49.9 & 57.5  \\
         Gemini-2.5-Pro & 63.5 & 57.3 & 62.0  \\
    \hline
    \rowcolor[RGB]{235, 245, 252}
    \multicolumn{4}{c}{\textcolor{black}{\textit{Open-Source Models}}}\\
    Qwen2.5-VL-7B & 43.3 & 45.6 & 43.8  \\
    Qwen2.5-VL-72B & 53.1 & 52.6 & 53.0  \\
    Qwen3-VL-8B & 52.1 & 52.4 & 52.2 \\
    InternVideo2.5-8B & 52.7 & 44.5 & 50.7  \\
    LLaVA-NeXT-Video-7B & 38.2 & 29.9 & 36.2  \\
    
    \hline
    \rowcolor[RGB]{235, 245, 252}
    \multicolumn{4}{c}{\textcolor{black}{\textit{3D Spatial Reasoning Models}}}\\
    VLM-3R (LLaVA-NeXT-Video-7B) & 36.9$_{\textcolor{ForestGreen}{\text{0.9}\downarrow}}$ & 24.7$_{\textcolor{ForestGreen}{\text{5.2}\downarrow}}$ & 33.9$_{\textcolor{ForestGreen}{\text{2.3}\downarrow}}$  \\
    VG-LLM (Qwen2.5-VL-7B) & 49.5$_{\textcolor{Maroon}{\text{6.2}\uparrow}}$   & 37.3$_{\textcolor{ForestGreen}{\text{8.3}\downarrow}}$ & 46.5$_{\textcolor{Maroon}{\text{2.7}\uparrow}}$  \\
    \hline
    \textbf{Our MLLM-4D (Qwen2.5-VL-7B)} & 59.4$_{\textcolor{Maroon}{\text{16.1}\uparrow}}$ & 49.7$_{\textcolor{Maroon}{\text{4.1}\uparrow}}$ & 57.0$_{\textcolor{Maroon}{\text{13.2}\uparrow}}$  \\
    \textbf{Our MLLM-4D (Qwen3-VL-8B)} & \textbf{63.1}$_{\textcolor{Maroon}{\text{11.0}\uparrow}}$ & \textbf{54.4}$_{\textcolor{Maroon}{\text{2.0}\uparrow}}$ & \textbf{61.0}$_{\textcolor{Maroon}{\text{8.8}\uparrow}}$  \\
	\bottomrule
	\end{tabular}
    }
        \caption{Evaluation on VLM4D benchmark.}
    \vspace{-0.6cm}
	\label{tab:vlm4d_bench}
\end{table}

\vspace{-1em}
\subsection{Comparisons on VLM4D Benchmark.} 
\vspace{-0.5em}
To demonstrate the generalizability of our model on out-of-distribution datasets, we further evaluate performance on VLM4D benchmark~\cite{zhou2025vlm4d}. As shown in Table~\ref{tab:vlm4d_bench}, MLLM-4D maintains superior performance, outperforming based model (Qwen3-VL-8B and Qwen2.5-VL-7B) and specialized 3D spatial reasoning models (VLM-3R and VG-LLM) by a significant margin.
Qualitative results provided in Fig.~\ref{fig: qa_example4} to Fig.~\ref{fig: qa_example6} illustrate that by leveraging the ST-CoT reasoning paradigm, the 4D spatiotemporal intelligence of MLLM-4D can effectively translate to diverse environments beyond our training distribution.

\vspace{-1em}
\subsection{Ablation Studies}
\vspace{-0.5em}

\begin{figure}[t] 
  \centering
  \begin{subfigure}[b]{0.23\textwidth}
    \centering
    \includegraphics[width=\textwidth]{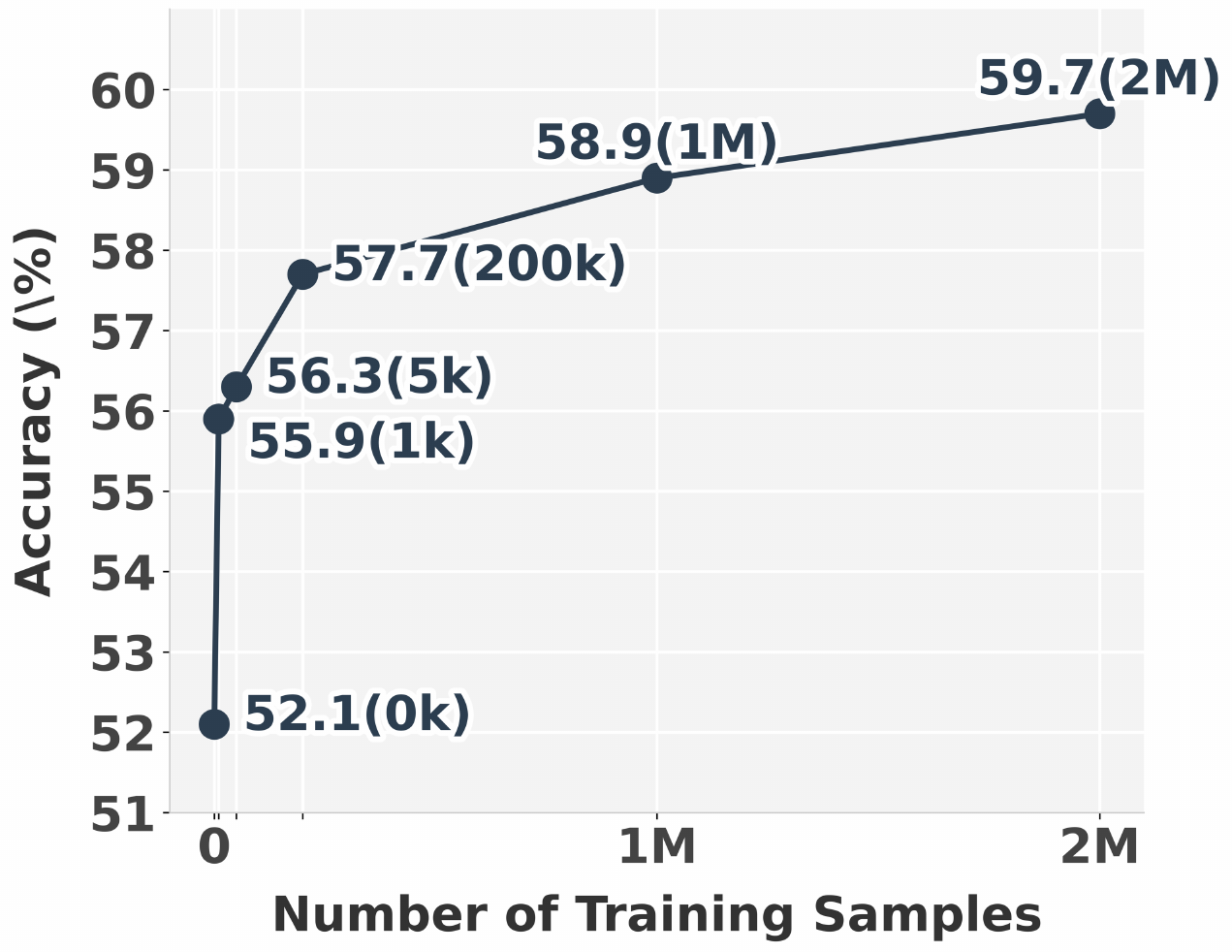}
    \caption{SFT training data scaling.}
  \end{subfigure}
  \vspace{2pt}
  \begin{subfigure}[b]{0.23\textwidth}
    \centering
    \includegraphics[width=\textwidth]{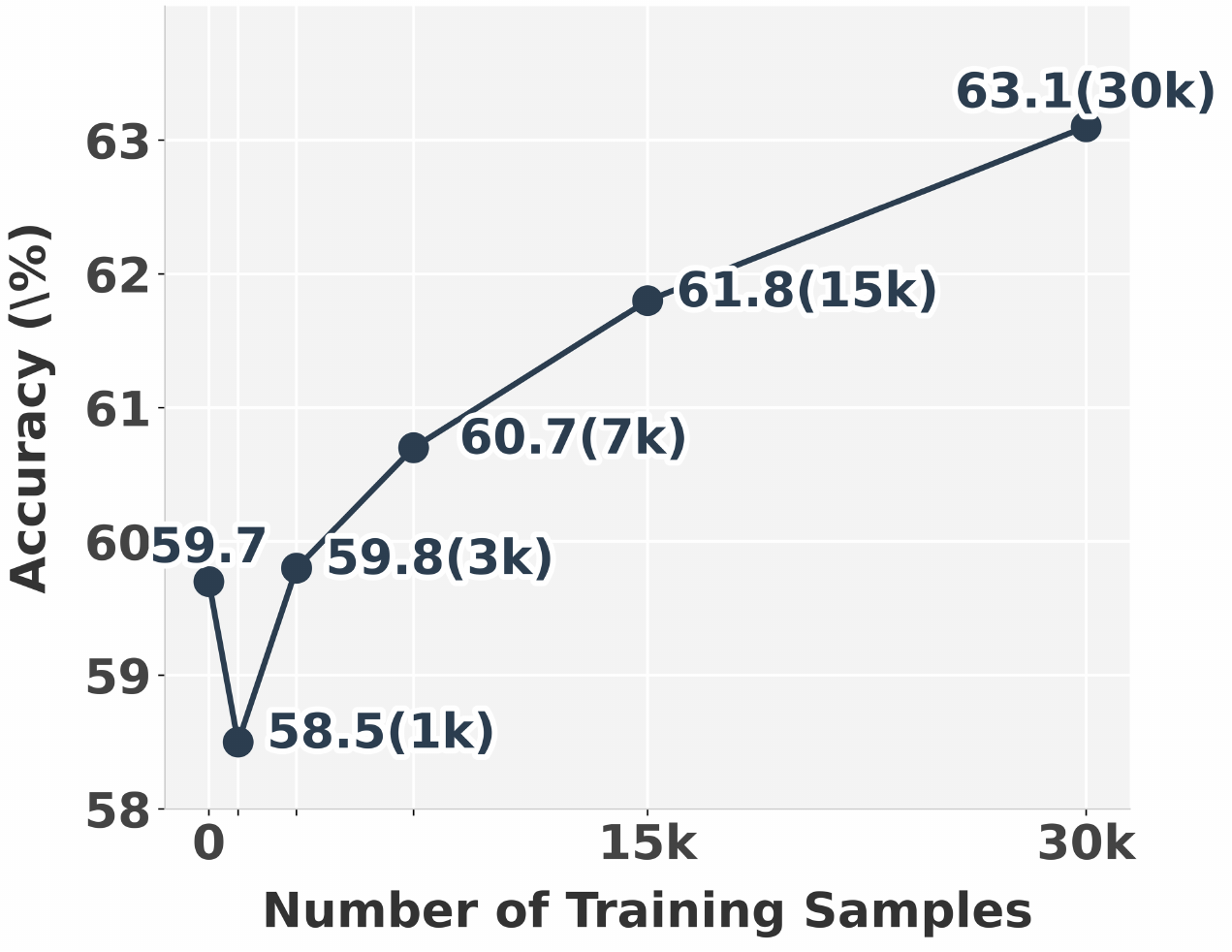}
    \caption{RFT training data scaling.}
  \end{subfigure}
  \vspace{-1em}
  \caption{Scalability of training data on SFT and RFT stage.}
  \label{fig: data scaling}
  \vspace{-2em}
\end{figure}

\noindent\textbf{Scalability of Training Data.} To evaluate the data scalability of MLLM4D-2M, we evaluate performance across training subsets of 10K, 50K, 200K, 1M, and 2M QA pairs using real-world videos of the VLM4D benchmark. As illustrated in Fig.~\ref{fig: data scaling} (a), the model exhibits a rapid performance surge in the initial scaling phase, with accuracy rising from a 52.1\% baseline to 57.7\% when using 200K samples. Beyond this threshold, performance continues to scale consistently, reaching 59.7\% at 2M samples. This sustained upward trajectory confirms that the MLLM4D-2M dataset effectively captures increasingly complex 4D spatial-temporal patterns as the data volume expands. We further analyze the scaling behavior during the RFT stage using subsets of MLLM4D-R1-30k ranging from 1K to 30K pairs. As shown in Fig.~\ref{fig: data scaling} (b), despite a minor initial fluctuation at 1K due to the model adapting to the reasoning format, the model exhibits a robust scaling trend beyond 3K samples. Accuracy improves from 59.8\% to a peak of 63.1\% at 30K samples, highlighting the scalability of our MLLM4D-R1-30k dataset.

\vspace{-0.2em}
\begin{table}[t]
	\centering
	\begin{tabular}{l|cc}
    \toprule
	Models & MLLM4D-Bench & VLM4D \\
	\hline
    Baseline & 35.3 & 52.1  \\
    SFT~(w/o data in Sec.~\ref{sec:scalable st data curation}) & 59.9 & 56.2  \\
    SFT~(w/ data in Sec.~\ref{sec:scalable st data curation})) & 70.1 & 59.7  \\
    GRPO~(w/o ST-reward) & 70.5 & 61.4 \\
    GRPO~(w/ ST-reward) & 72.7 & 63.1 \\
    \bottomrule
	\end{tabular}
    \caption{Ablation study of MLLM-4D framework.}
    \vspace{-3em}
	\label{tab:ablation_1}
\end{table}
\noindent\textbf{Effectiveness of MLLM-4D.} Based on the ablation study presented in Table~\ref{tab:ablation_1}, our post-training framework progressively enhances 4D spatiotemporal reasoning capabilities of MLLM across both the MLLM4D-Bench and VLM4D benchmarks. The results highlight the following key insights: (1) \textit{Impact of SFT}: the transition from the Baseline to SFT version of MLLM-4D yields a substantial performance leap, particularly on the MLLM4D-Bench, where scores nearly double from 35.3\% to 70.1\%. This confirms that the MLLM4D-2M dataset effectively establishes a robust foundational 4D understanding. (2) \textit{Scalable Spatial-temporal Data Curation}: we compared our primary data curation pipeline against an alternative method based on monocular videos (see details in Appendix~\ref{appendix: Details of data curation pipeline based on monocular videos}). While both large-scale 4D datasets drive significant improvements, the pipeline proposed in Sec.~\ref{sec:scalable st data curation} yields superior results. This suggests that our pipeline provides higher-quality data, whereas monocular-based pipelines often suffer from depth ambiguity and diminished spatial accuracy. (3) \textit{Benefits of RFT}: applying GRPO with standard accuracy and format rewards provides robust performance gains over SFT. (4) \textit{Advantage of ST-reward}: the full post-training, including the ST-reward, achieves the highest performance (72.7\% on MLLM4D-Bench and 63.1\% on VLM4D). This validates that grounding the reasoning process in 4D physical quantities through ST-reward functions significantly boosts the model's capacity for complex spatiotemporal reasoning.

\vspace{-1.1em}
\section{Conclusion}
\vspace{-0.5em}

We present MLLM-4D, a comprehensive framework designed to advance the 4D spatiotemporal reasoning capabilities of MLLMs. To address the scarcity of high-quality data, we introduce an automated curation pipeline for large-scale 4D instructional pairs. We bridge the gap in specialized 4D-aware modeling by establishing foundational 4D understanding via SFT and subsequently unlocking advanced 4D reasoning capabilities by employing GRPO with specialized ST-CoT and ST-reward functions. We hope that our data, models, and methodology inspire future research into 4D spatiotemporal intelligence and facilitate the development of interactive AI systems in the real world.

\section{Impact Statement}
MLLM-4D provides a comprehensive framework designed to boost the spatial-temporal intelligence of MLLMs. This advancement is particularly beneficial for interactive AI systems such as VR/AR, autonomous driving, and robotics. In varied application scenarios, it is essential to follow the corresponding usage guidelines to ensure its proper and ethical application, minimizing any potential risks.  


\bibliography{example_paper}
\bibliographystyle{icml2026}

\newpage
\clearpage
\section{Appendix}
\subsection{Implementation Details}
\noindent\textbf{Training Details.} MLLM-4D is built on Qwen3-VL-8B-Instruct~\cite{Qwen3-VL} and Qwen2.5-VL-7B-Instruct~\cite{Qwen2.5-VL}. During training, we utilize LoRA~\cite{hulora} configured with an update matrix rank of 128 and an adaptation scaling parameter of 256, and we limit video frames to 32. In the SFT stage, we train the model using Adam optimizer for one epoch ($30k$ steps). We employ a cosine learning rate schedule with a peak learning rate of $2\times10^{-5}$, a warmup ratio of 0.01, and a global batch size of 8. In the cold start stage, we use a similar setting as in the SFT stage to train the model for about 200 steps. In the RFT stage, we perform 12 rollouts per question and set the default sampling temperature to 1. The $\lambda_{Acc}, \lambda_{Fmt}, \lambda_{ST}, \lambda_{1}, \lambda_{2}, \lambda_{Cam}, \lambda_{Obj}$ are set to 0.5, 0.2, 0.3, 0.5, 0.5, 0.5, 0.5, respectively.  We train the model for $15k$ steps with a KL divergence coefficient $\beta$ of 0.1 and a learning rate of $5\times10^{-5}$. All experiments were conducted on 8 H100 80G GPUs; the training takes 12 hours for the SFT and cold start stage, and 50 hours for the RFT stage.

\noindent\textbf{Comparison Baselines.} We compare our MLLM-4D with a comprehensive suite of state-of-the-art MLLMs. For proprietary models, we include GPT-4o~\cite{hurst2024gpt}, Gemini2.5-Flash~\cite{comanici2025gemini} and Gemini2.5-Pro~\cite{comanici2025gemini}. Regarding open-source MLLMs, we consider various scales of Qwen2.5-VL~\cite{Qwen2.5-VL}, Qwen3-VL~\cite{Qwen3-VL}, LLaVA-NeXT-Video~\cite{zhang2024llavanext-video}, InternVideo2.5~\cite{wang2025internvideo2}, InternVL2.5~\cite{chen2024expanding}, and InternVL3.5~\cite{wang2025internvl3} series. We also include specialized 3D spatial reasoning models, such as VG-LLM~\cite{zheng2025learning} and VLM-3R~\cite{fan2025vlm}.

\subsection{Details of MLLM4D-2M and MLLM4D-R1-30k Dataset Construction} \label{appendix: Details of MLLM4D-2M}

\begin{figure}[t]
  \centering
   \includegraphics[width=\linewidth]{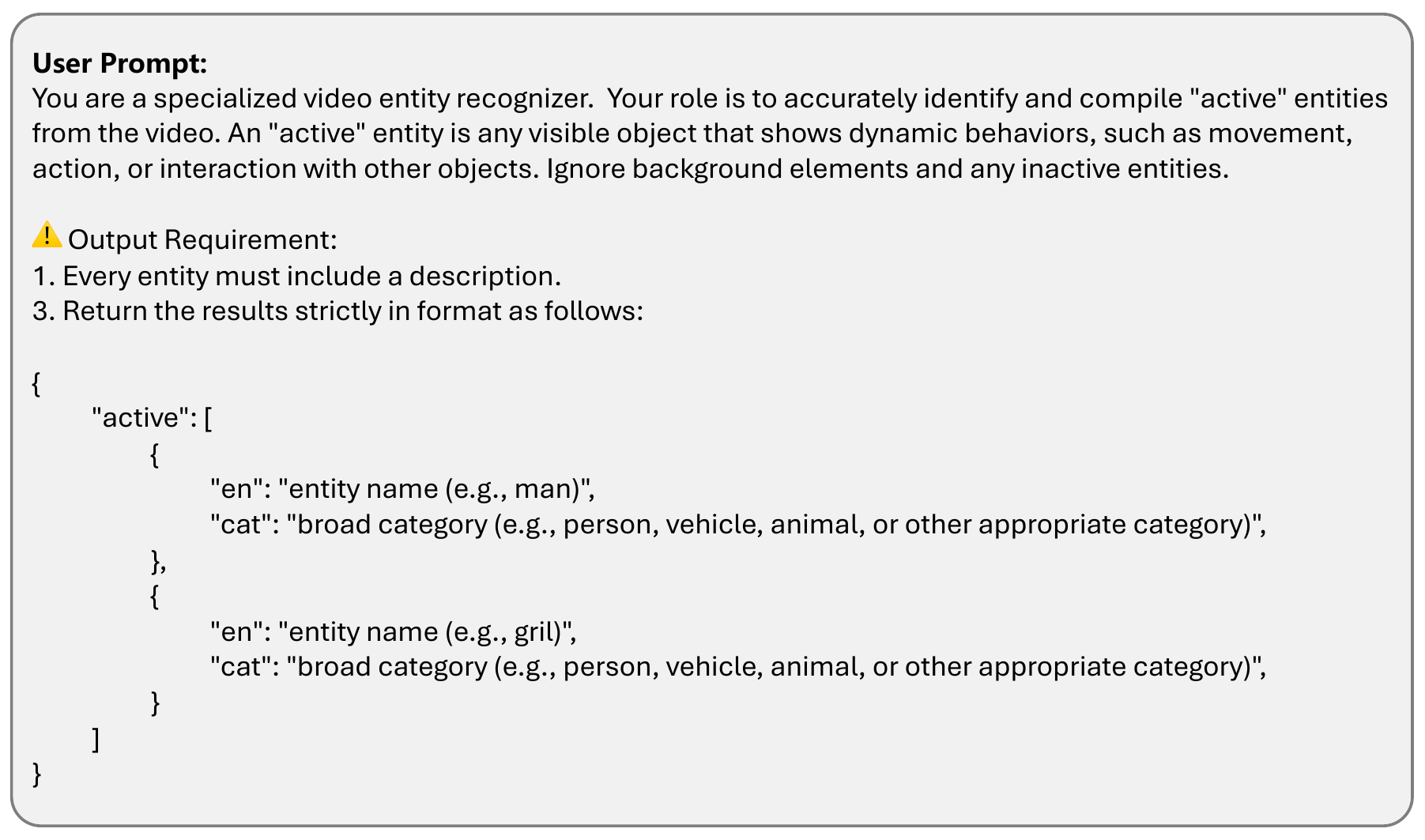}
   \caption{Illustration of the prompt used to identify all moving entities and extract their corresponding noun categories.}
   \label{fig: moving_object_prompt}
\end{figure}

\noindent\textbf{Moving Object Noun categories extraction.} Given the video sequence $\{I_i\}_{i=1}^{K}$, we employ a video MLLM, \emph{i.e.} Gemini-2.5-flash~\cite{comanici2025gemini} using the prompt shown in Fig.~\ref{fig: moving_object_prompt}, to identify all moving entities and extract their corresponding noun categories.

\noindent\textbf{QA Pair Generation.} We organize the extracted information into a structured meta-data format. This encompasses per-frame camera poses $\{C_i\}_{i=1}^{K}$, where $C_i = [R_i | t_i]$ consists of a $3\times3$ rotation matrix $R_i$ and a $3\times1$ translation vector $t_i$; object-level metric 3D points $\{P_{i,m}\}_{i=1,...,K; m=1,...,M}$; and their corresponding semantic descriptions $\{T_m\}_{m=1}^{M}$.

Using \textbf{\textit{physical-based spatial-temporal computation}}, we subsequently generate QA pairs of different tasks across several spatiotemporal reasoning tasks:
\begin{itemize}[leftmargin=2em]
    \item \textit{Camera Absolute Distance}: Measures the movement of camera between any two frames $i$ and $j$. We first compute the camera center in world coordinates as $Center_i = -R^T t$. Then the absolute distance $d_{cam\_abs\_dis}$ is the $L_2$ norm: $d_{cam\_abs\_dis} = \| Center_j - Center_i \|_2 = \sqrt{\sum_{k=1}^{3} (Center_{j,k} - Center_{i,k})^2}$. Question template: “Approximately how far (in meters) did the camera move between $<\texttt{frame}\ i>$ and $<\texttt{frame}\ j>$?”

    \item \textit{Camera Relative Direction}: Determines movement relative to the camera’s orientation at frame $i$ between any two frames $i$ and $j$. we first compute the the world-space displacement $\Delta t = t_j - t_i$. To determine the movement relative to the camera's perspective at frame $i$, we project this vector into the local camera coordinate system using the transpose of the rotation matrix $R_i^T$: $d_{cam\_rel\_dir} = R_i^T \cdot \Delta t$. The resulting vector $d_{cam\_rel\_dir} = [dx, dy, dz]$ is interpreted via standard camera conventions, where $+X, +Y, +Z$ correspond to right, down, and forward direction, respectively. Question template: “During the sequence between $<\texttt{frame}\ i>$ and $<\texttt{frame}\ j>$, what was the primary consistent translation of the camera's movement relative to its position at the start?” 

    \item \textit{Object Absolute Distance}: Computes the movement distance of an object between any two frames $i$ and $j$. We define the object centroid at frame $i$ as $\bar{P}_i = \frac{1}{M} \sum_{m=1}^{M} P_{i,m}$. Then, the object absolute distance $d_{obj\_abs\_dis}$ is calculated as the $L_2$ norm of the vector connecting their centroids: $d_{obj\_abs\_dis} = \| \bar{P}_j - \bar{P}_i \|_2 = \sqrt{\sum_{k=1}^{3} (\bar{P}_{j,k} - \bar{P}_{i,k})^2}$.
    Question template: “Approximately how far (in meters) did $<object\ m\ description>$ move between $<\texttt{frame}\ i>$ and $<\texttt{frame}\ j>$?” 
    
    \item \textit{Object-Camera Absolute Distance}: Computes the proximity of an object at frame $j$ to the camera at frame $i$. We first compute the camera center in world coordinates as $Center_i = -R^T t$. Then the absolute distance $d_{\text{obj}}$ between the camera and the object is defined as the minimum distance from the camera center to any point within the object's point set: $d_{\text{obj-cam\_abs}} = \min\| P_{j,m} - Center_i \|_2$. Question template: “What is the approximate distance (in meters) between the camera (or the observer filming) in $<\texttt{frame}\ i>$ and the nearest point of the $<\texttt{object}\ m\ \texttt{description}>$ in $<frame\ j>$?”
    
    \item \textit{Object-Camera Relative Distance}: Determines whether the camera and object are converging or diverging between any two frames $i$ and $j$. We first find the camera center to any point within the object's point set: $d_i = \min\| P_{i,m} - Center_i \|_2$. Then the relative change in distance, $\Delta d$, is the difference between the instantaneous distances at frame $j$ and frame $i$: $\Delta d = d_j - d_i$. Based on a distance threshold $\tau$, the relative movement is classified into a discrete state $S = \text{farther if } \Delta d > \tau; \text{closer if } \Delta d < -\tau; \text{otherwise not moving}$.
    Question template: “During the sequence between $<\texttt{frame}\ i>$ and $<\texttt{frame}\ j>$, is the distance between $<\texttt{object}\ m\ \texttt{description}>$ and the camera (or the observer filming) getting closer or farther away?” 

    \item \textit{Object-Camera Relative Direction}: determine the relative direction between the camera and the moving object between is getting left/right or closer/farther between any two frames $i$ and $j$. We first transform the object points from both time steps into the camera's local coordinate system at frame $i$: the local camera coordinates: $P_{i}^c = R_i P_i + t_i$. Then we calculate the centroid of the object in the camera's coordinate space for both sets:$\bar{P}_i^c = \frac{1}{M} \sum_{n=1}^{M} (R_i P_{i,m} + t_i), \quad \bar{P}_j^c = \frac{1}{M} \sum_{m=1}^{M} (R_i P_{j,m} + t_i)$. We then classify the lateral change $\Delta x = \bar{P}_{j,x}^c - \bar{P}_{i,x}^c$ (left/right) and the longitudinal change $\Delta z = \bar{P}_{j,z}^c - \bar{P}_{i,z}^c$ (closer/farther) using threshold $\tau$. Question template: “During the sequence between $<\texttt{frame}\ i>$ and $<\texttt{frame}\ j>$, is $<\texttt{object}\ m\ \texttt{description}>$ getting left or right from the camera (or the observer filming) relative to camera's position at the start?” or “During the sequence between $<\texttt{frame}\ i>$ and $<\texttt{frame}\ j>$, is $<\texttt{object}\ m\ \texttt{description}>$ getting closer or farther away from the camera (or the observer filming) relative to camera's position at the start?”
\end{itemize}

\noindent\textbf{Data Filtering and Balancing Protocols.} After calculating the ground-truth values, we apply templates to formulate the final QA pairs. We implement several filtering and balancing protocols to ensure quality: we limit the number of QA pairs per video to maintain scene diversity and shuffle multiple-choice options to eliminate positional bias. Additionally, the distractors for numerical options are randomly generated within $25\%$--$175\%$ of the true value to prevent unrealistic shifts. After filtering, we retain $2M$ high-quality QA pairs across approximately $100k$ videos, forming the large-scale \textit{MLLM4D-2M} dataset for supervised fine-tuning. 

\begin{figure}[t]
  \centering
   \includegraphics[width=\linewidth]{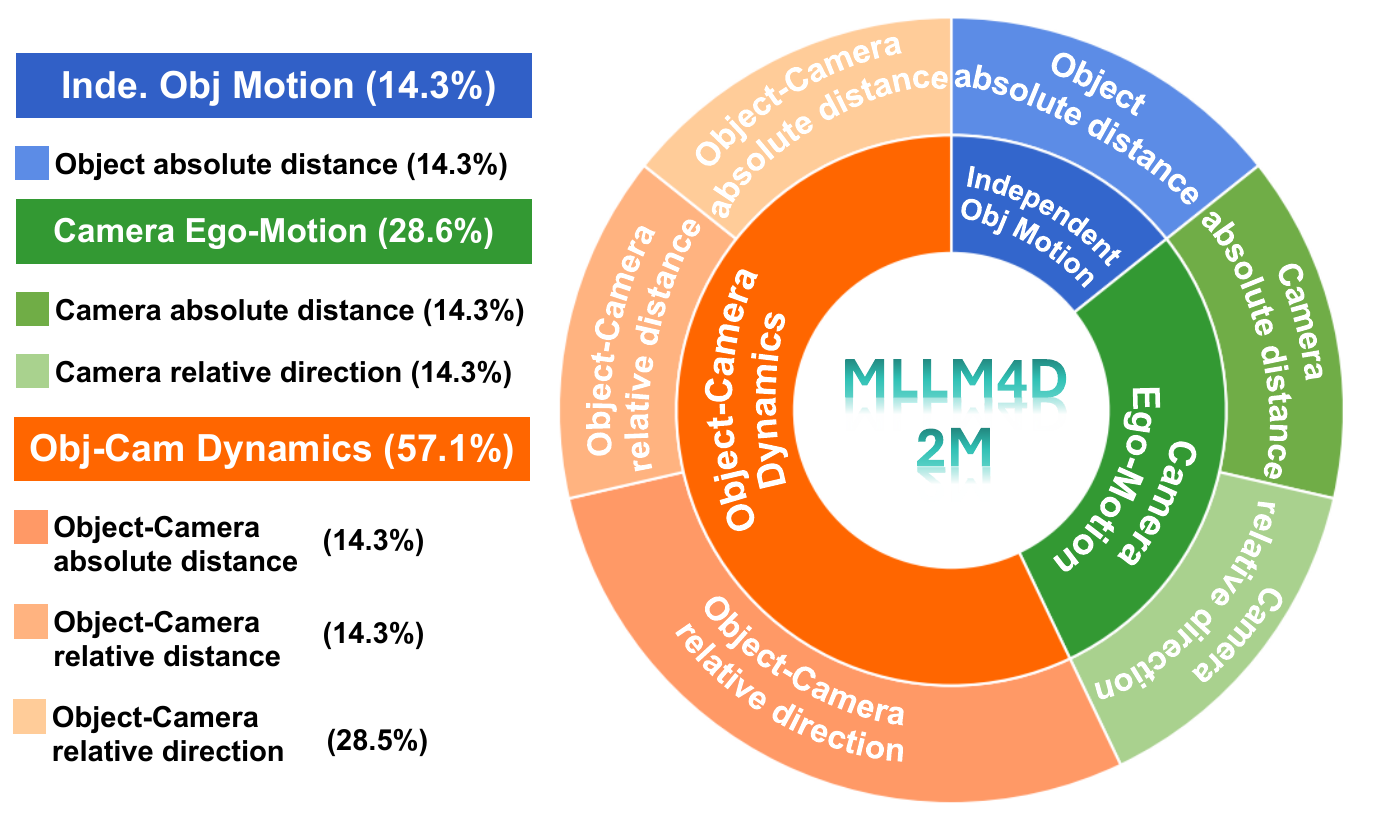}
   \caption{The components of our MLLM4D-2M.}
   \label{fig: figure2_MLLM4D-2M}
\end{figure}

\begin{figure}[t]
  \centering
   \includegraphics[width=\linewidth]{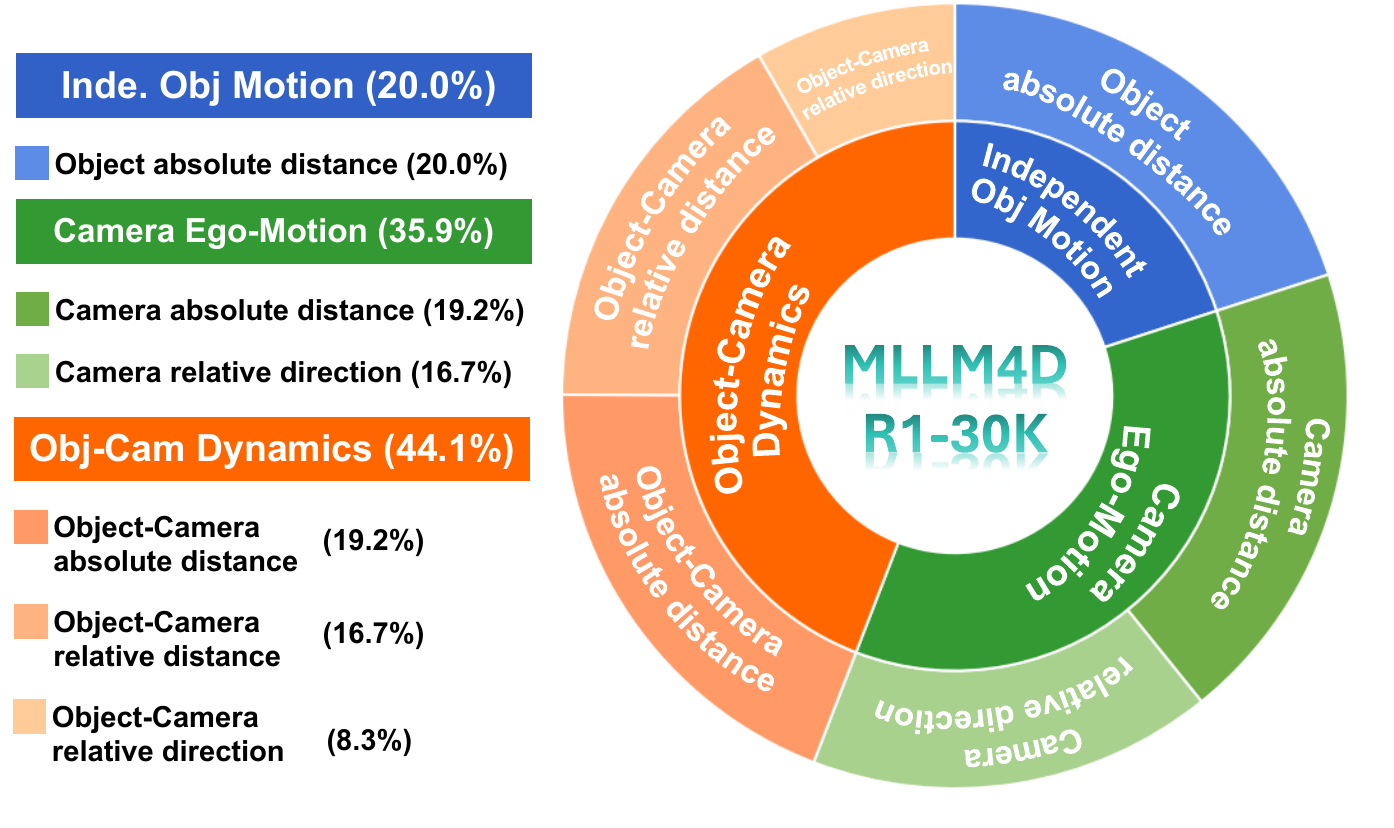}
   \caption{The components of our MLLM4D-R1-30k.}
   \label{fig: figure2_MLLM4D-R1-30k}
\end{figure}

\begin{figure}[t]
  \centering
   \includegraphics[width=\linewidth]{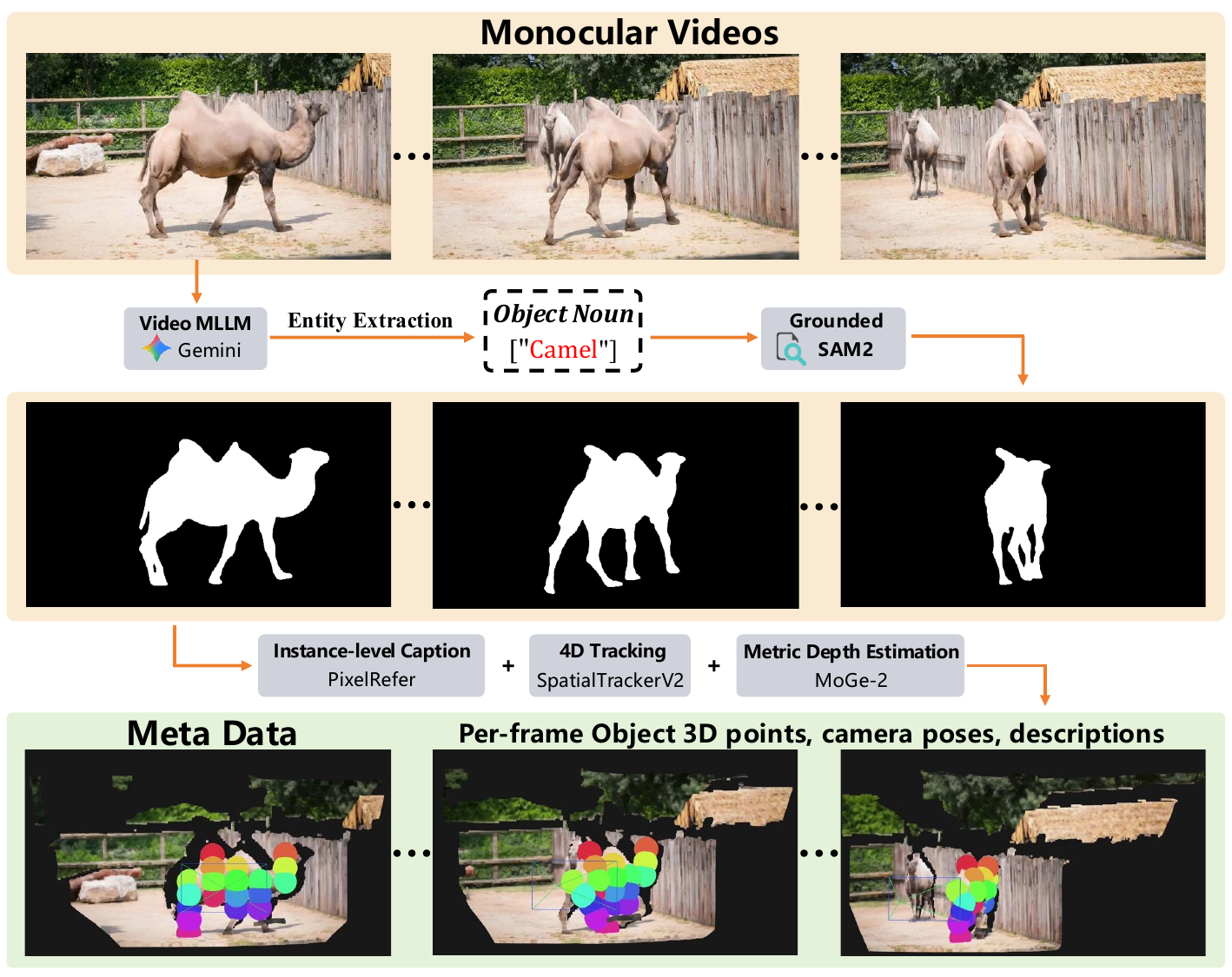}
   \caption{An alternative data pipeline based on monocular videos.}
   \label{fig: monocular pipeline}
\end{figure}

\subsection{Details of Cold Start}\label{sec: cold_start details}
To align the model's output with the desired reasoning format, we conducted a brief cold-start fine-tuning phase consisting of 200 steps before GRPO training, following the same hyperparameters as SFT. 
The cornerstone of this phase involves constructing a reasoning dataset with Chain-of-Thought (CoT) annotations derived from pre-collected question-answer pairs. The construction process is detailed as follows:

\noindent\textbf{Subset Sampling.} We begin by sampling a subset $\mathcal{D}_0 = \bigcup_{s=1}^{7} \mathcal{D}_s$ from the MLLM4D-2M dataset, which is constructed by drawing samples across all $7$ distinct scenarios covered in the original dataset. 
Specifically, $\mathcal{D}_s = \{ \mathcal{I}_s^{i} \}_{i=1}^{N}= \{ \langle \mathcal{Q}_s^{i}, \mathcal{A}_s^{i}, \mathcal{V}_s^{i} \rangle \}_{i=1}^{N}$, where each instance is uniformly sourced from the diverse scenario pool.

\noindent\textbf{Multi-path CoT Generation.} For each sample $\mathcal{I}_s^{i}\in \mathcal{D}_0$, we utilize Gemini2.5-Pro \cite{comanici2025gemini} to generate K independent reasoning processes $\mathcal{\hat{T}}_s^{i,k}$ and corresponding answers $\mathcal{\hat{A}}_s^{i,k}$.
We then compute the reward $r_s^{i,k}=\text{Reward}(\mathcal{\hat{A}}_s^{i,k},\mathcal{A}_s^{i})$ for each reasoning-answer pair, where $\text{Reward}(.,.)$ is the reward function described in Sec \ref{sec:rft}.  Consequently, we obtain a set of outputs $\mathcal{O}_{s}^{i}=\{ \langle \mathcal{\hat{T}}_s^{i,k}, \mathcal{\hat{A}}_s^{i,k}, r_s^{i,k} \rangle \}_{k=1}^{K}$ for each $\mathcal{I}_s^{i}\in \mathcal{D}_0$.

\begin{figure*}[t]
    \centering
    \resizebox{\textwidth}{!}{
        \includegraphics{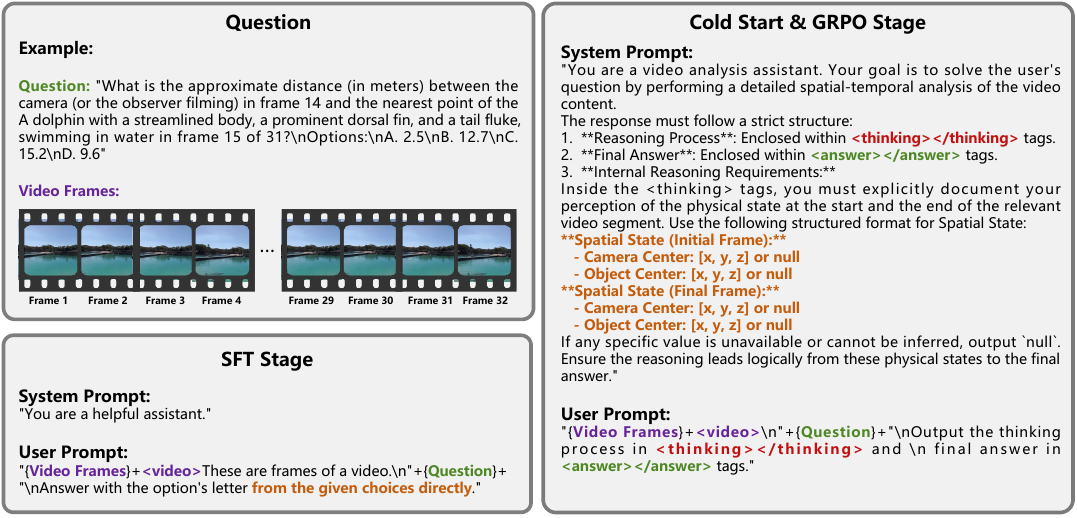} 
    }
    \caption{Detailed system prompt and user prompt setting for our SFT, Cold Start and GRPO stage.}
    \label{fig:prompt_setting}
\end{figure*}

\begin{figure*}[t]
    \centering
    \resizebox{\textwidth}{!}{
        \includegraphics{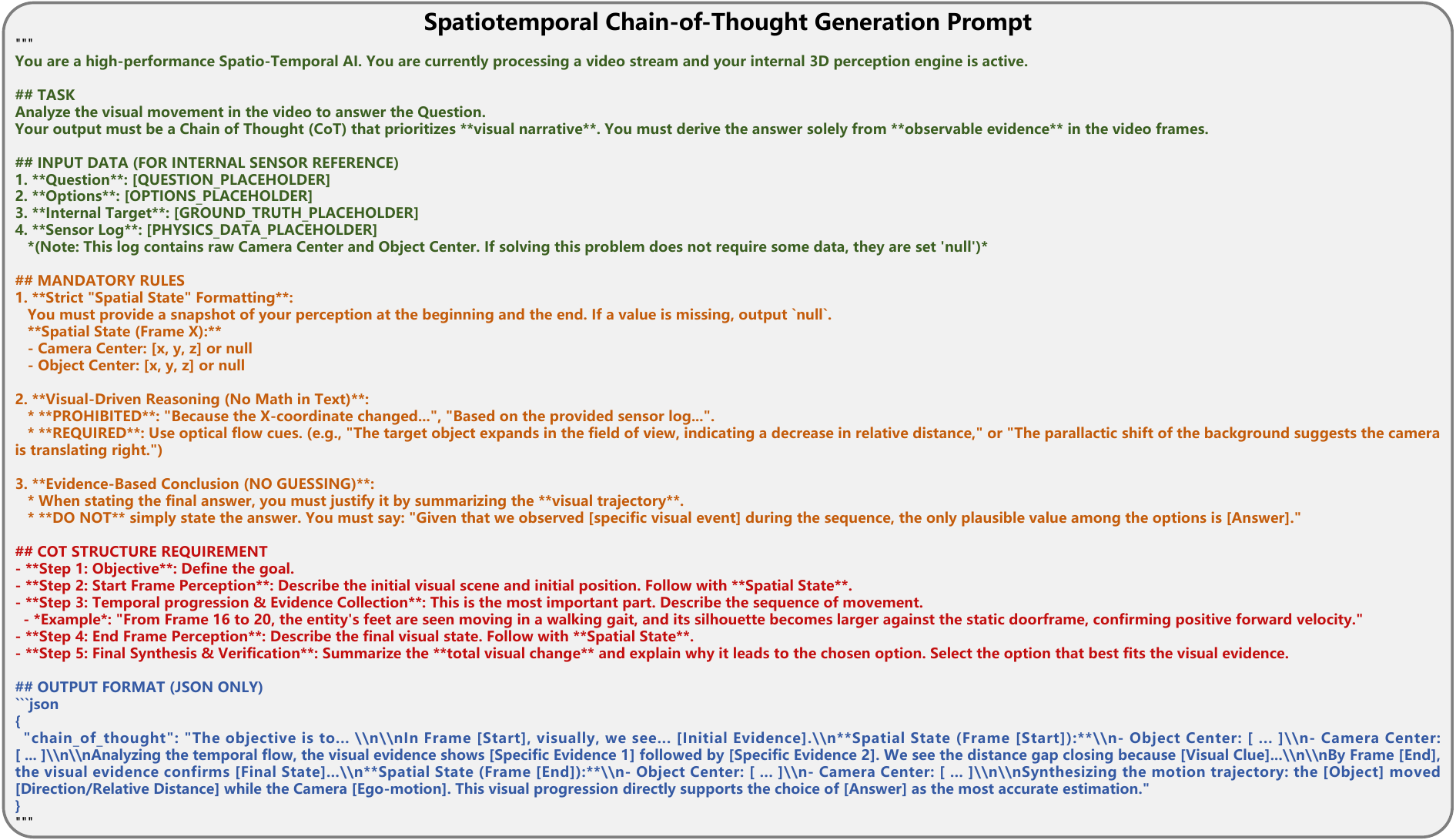} 
    }
    \caption{Detailed structure of the Spatiotemporal Chain of Thought (ST-CoT) Generation Prompt. The prompt is color-coded into four functional modules: \textcolor[HTML]{4F7928}{$\blacksquare$} \textbf{System Role \& Task} defines the Al persona and core mission; \textcolor[HTML]{c25500}{$\blacksquare$} \textbf{Input Data \& Mandatory Rules} enforces visual-driven reasoning and constraints; \textcolor[HTML]{A8201A}{$\blacksquare$} \textbf{CoT Structure} provides the step-by-step requirements for the reasoning chain; \textcolor[HTML]{1F5C7A}{$\blacksquare$} \textbf{Output Format} specifies the data placeholders and JSON schema.}
    \label{fig: ST-CoT prompt}
\end{figure*}

\begin{figure*}[t]
    \centering
    \resizebox{\textwidth}{!}{
        \includegraphics{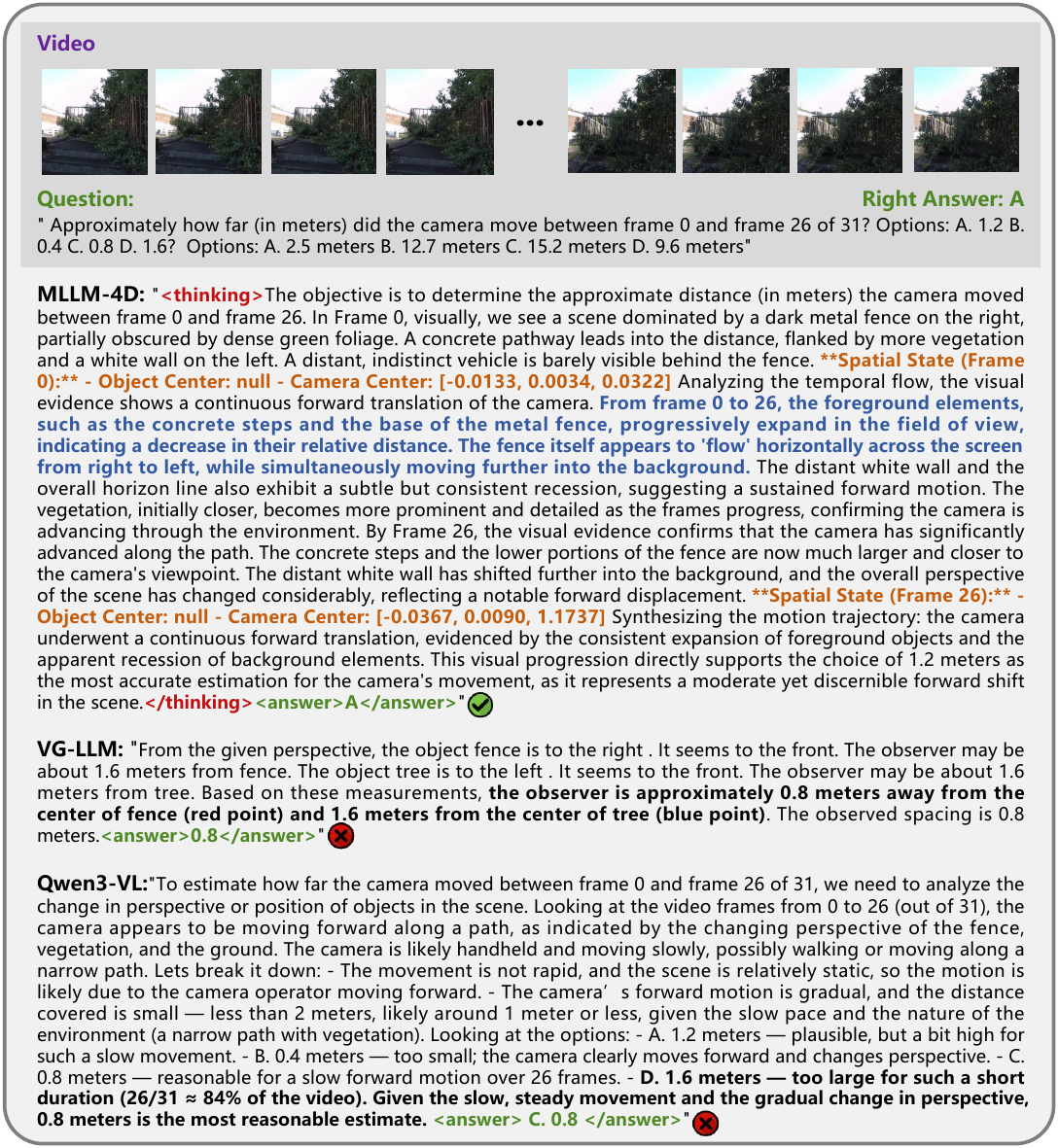} 
    }
    \caption{Qualitative comparison on MLLM4D-Bench.}
    \label{fig: qa_example1}
\end{figure*}

\begin{figure*}[t]
    \centering
    \resizebox{\textwidth}{!}{
        \includegraphics{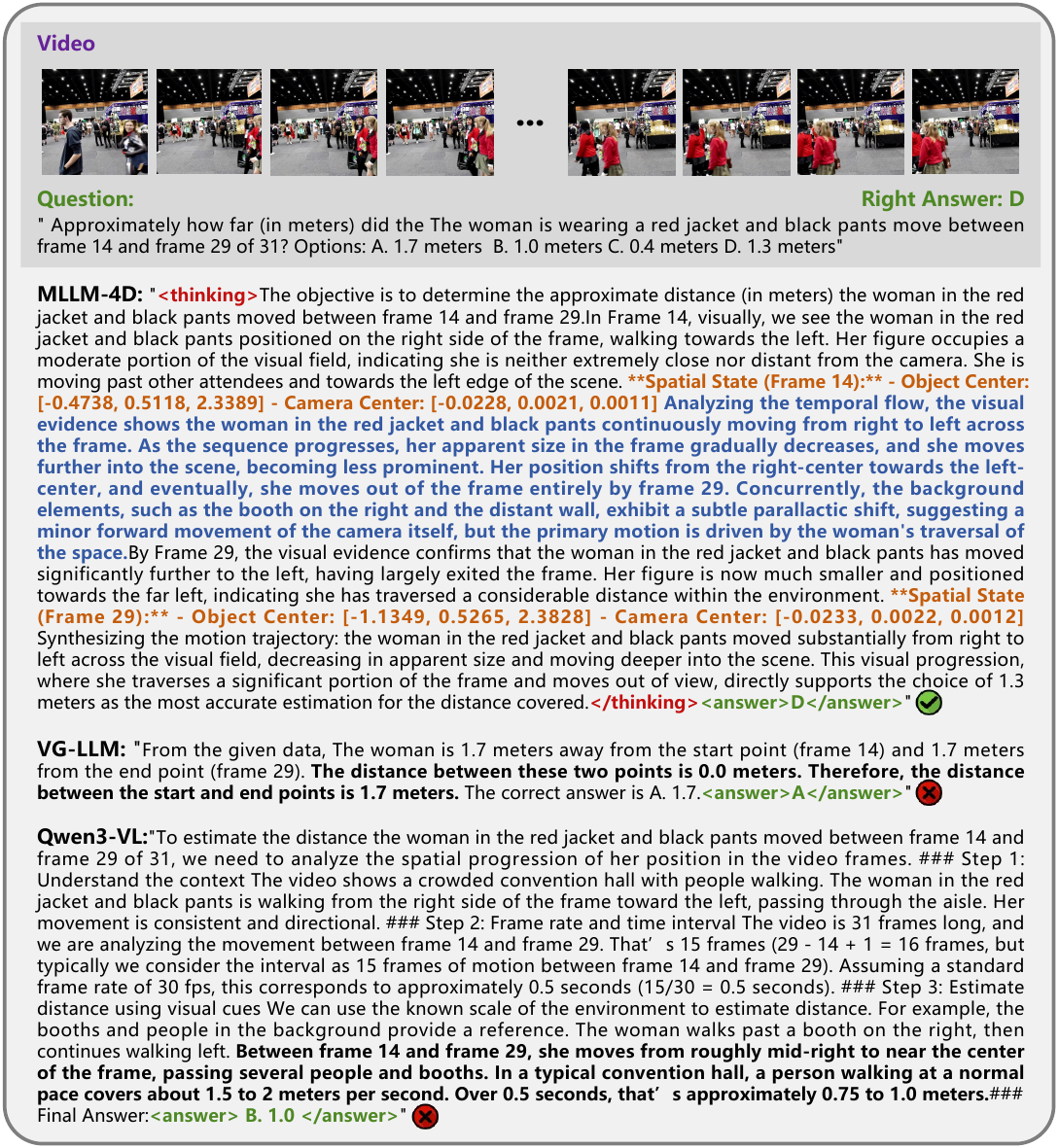} 
    }
    \caption{Qualitative comparison on MLLM4D-Bench.}
    \label{fig: qa_example2}
\end{figure*}

\begin{figure*}[t]
    \centering
    \resizebox{\textwidth}{!}{
        \includegraphics{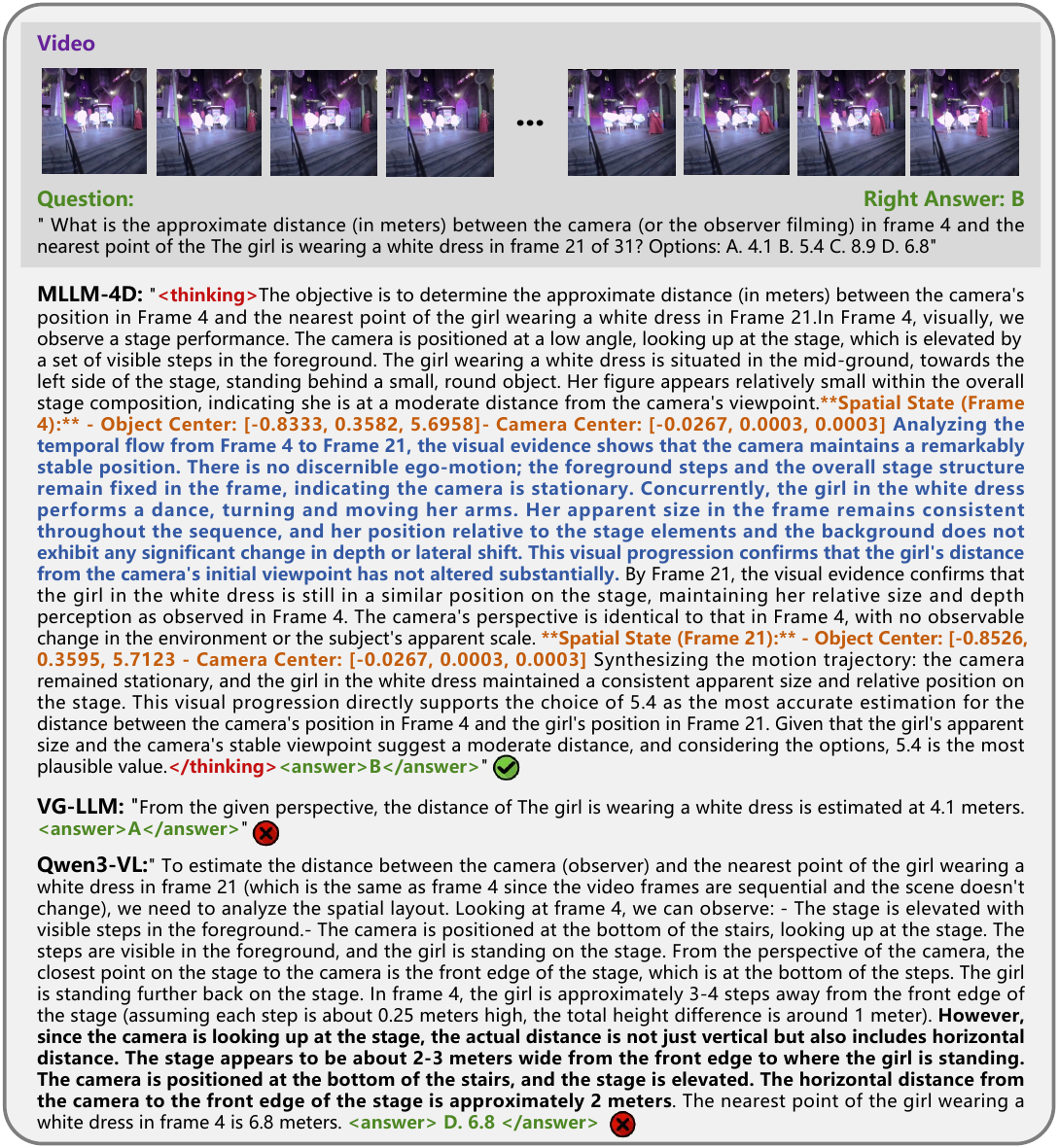} 
    }
    \caption{Qualitative comparison on MLLM4D-Bench.}
    \label{fig: qa_example3}
\end{figure*}

\begin{figure*}[t]
    \centering
    \resizebox{\textwidth}{!}{
        \includegraphics{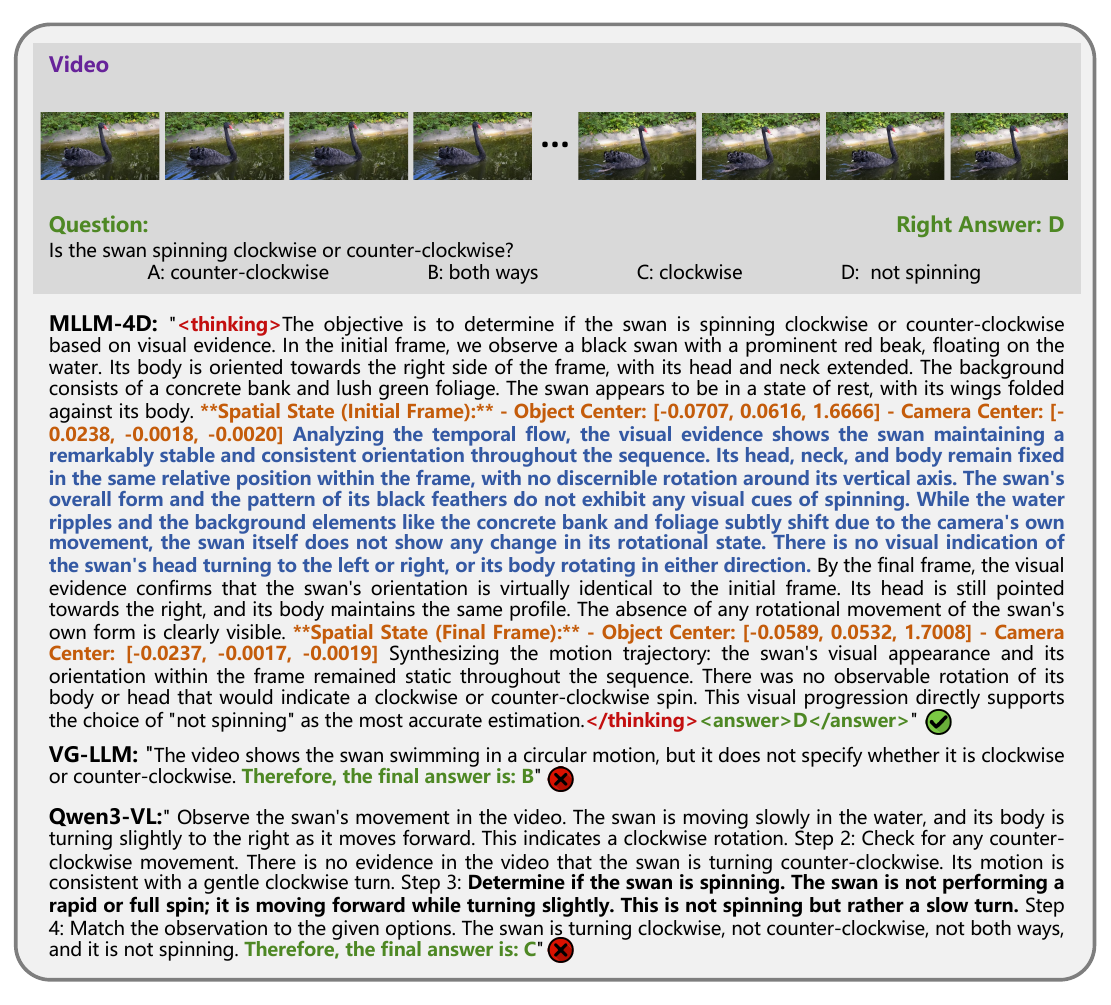} 
    }
    \caption{Qualitative comparison on VLM4D benchmark.}
    \label{fig: qa_example4}
\end{figure*}

\begin{figure*}[t]
    \centering
    \resizebox{\textwidth}{!}{
        \includegraphics{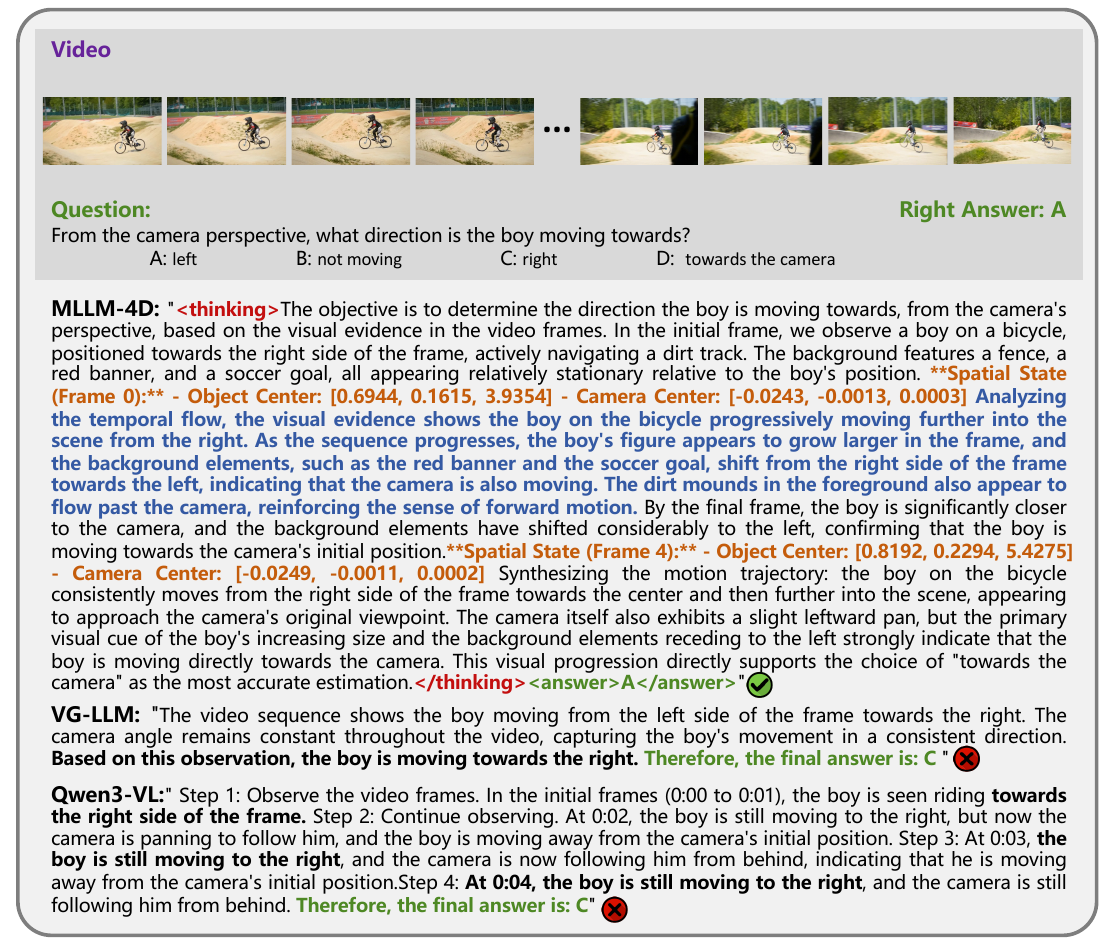} 
    }
    \caption{Qualitative comparison on VLM4D benchmark.}
    \label{fig: qa_example5}
\end{figure*}

\begin{figure*}[t]
    \centering
    \resizebox{\textwidth}{!}{
        \includegraphics{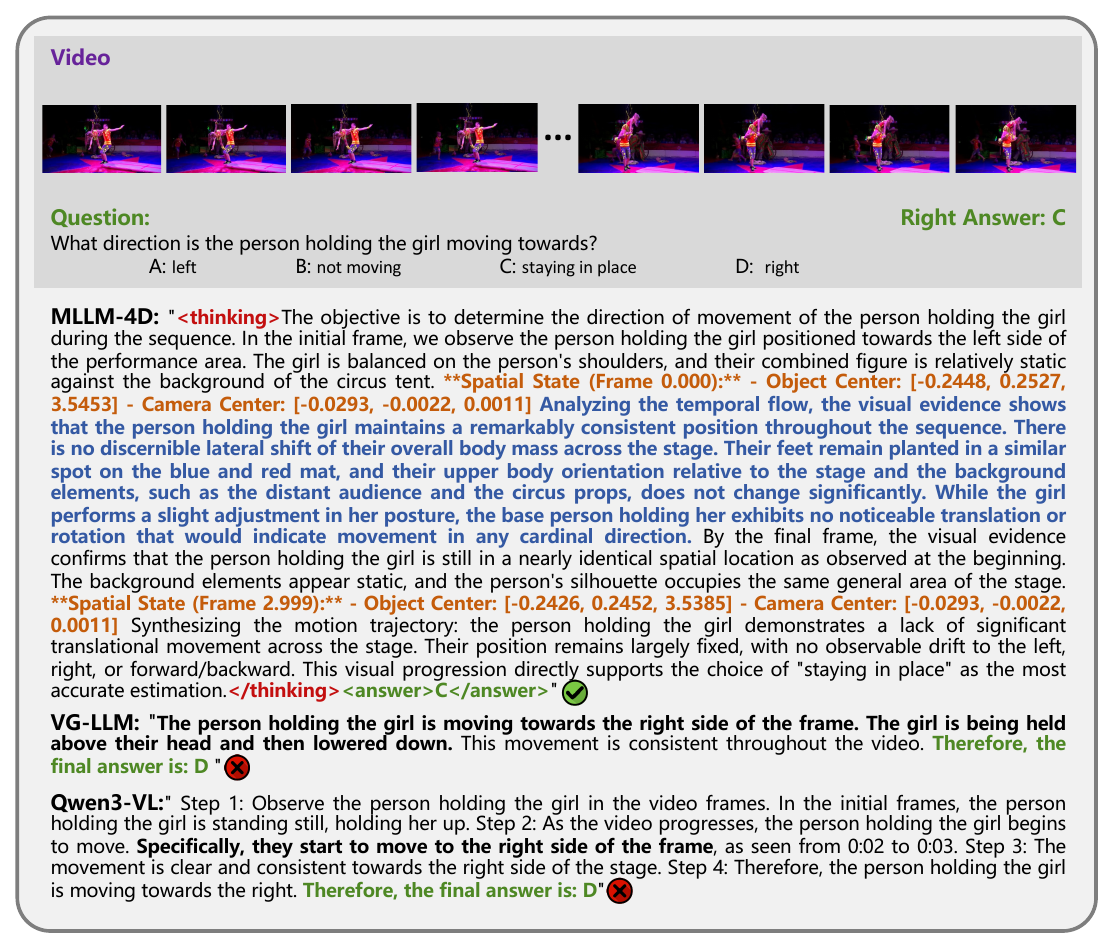} 
    }
    \caption{Qualitative comparison on VLM4D benchmark.}
    \label{fig: qa_example6}
\end{figure*}

\noindent\textbf{Scenario-specific Filtering.} Since the generative quality of Gemini2.5-Pro \cite{comanici2025gemini} may vary across different data distributions, applying a global reward threshold can lead to an imbalance across scenarios. To mitigate this, we adopt a scenario-specific filtering strategy. For each sample $\mathcal{I}_s^{i} \in \mathcal{D}_0$, we first identify the output with the highest reward, denoted as $\hat{r}_s^i = \arg\max_k r_s^{i,k}$. We then compute a scenario-dependent threshold $\tau_s$ by averaging the maximum rewards within each scenario:
\begin{equation}
\tau_s = \frac{1}{N} \sum_{i=1}^{N} \hat{r}_s^i, \quad s \in {1, \dots, 7}.
\end{equation}
An item $\mathcal{I}_s^i$ is preserved in the final cold-start set if and only if $\hat{r}_s^i \ge \tau_s$. This strategy ensures that the model learns from the top-performing generations relative to each scenario's complexity, maintaining a balanced representation of all $7$ scenarios in the reasoning dataset.
This rule preserves approximately the top 50\% of generations per question type while discarding degenerate (zero-reward) outputs. In practice, we set N = 2000 and K = 3, and finally, we get 7000 samples in the cold start set. We provide a pseudocode for this process in Algorithm \ref{alg:cold_start}:

\noindent\textbf{Other Details.} Figure~\ref{fig:prompt_setting} presents the prompts used in the SFT, Cold Start, and GRPO stages. For the SFT stage,
we adopt the default system prompt of Qwen3-VL \cite{Qwen3-VL}, namely, “You are a helpful assistant.” In the Cold Start and GRPO stage, we use specially designed system prompts to guide the model’s output of 4D information.

\begin{algorithm}[t]
\caption{Scenario-Adaptive Cold-Start Construction}
\label{alg:cold_start}
\begin{algorithmic}[1] 
\STATE \textbf{Input:} Initial subset $\mathcal{D}_0 = \bigcup_{s=1}^7 \mathcal{D}_s$, Model $\mathcal{M}$, Reward function $R(\cdot)$
\STATE \textbf{Output:} Filtered cold-start dataset $\mathcal{D}_{cold}$
\STATE $\mathcal{D}_{cold} \leftarrow \emptyset$

\FOR{each scenario $s \in \{1, \dots, 7\}$}
    \STATE \COMMENT{Step 1: Multi-path Generation}
    \FOR{each sample $\mathcal{I}_s^{i} \in \mathcal{D}_s$}
        \STATE Get $\text{K}$ paths: $\{ \langle \hat{\mathcal{T}}_s^{i,k}, \hat{\mathcal{A}}_s^{i,k}, r_s^{i,k} \rangle \}_{k=1}^{K}$ using $\mathcal{M}$
    \ENDFOR
    
    \STATE \COMMENT{Step 2: Scenario-specific Filtering}
    \STATE Compute scenario mean reward as threshold: \\$\tau_s \leftarrow \frac{1}{N } \sum_{i=1}^{N} \hat{r}_s^i=\frac{1}{N } \sum_{i=1}^{N}\arg\max_k r_s^{i,k}$
    
    \FOR{each sample $i$ and path $k$ in scenario $s$}
        \IF{$r_s^{i,k} \geq \tau_s$ \textbf{and} $r_s^{i,k} > 0$}
            \STATE $\mathcal{D}_{cold} \leftarrow \mathcal{D}_{cold} \cup \{ \langle \mathcal{I}_s^i, \hat{\mathcal{T}}_s^{i,k}, \hat{\mathcal{A}}_s^{i,k} \rangle \}$
        \ENDIF
    \ENDFOR
\ENDFOR

\STATE \textbf{Return:} $\mathcal{D}_{cold}$
\end{algorithmic}
\end{algorithm}

\subsection{Details of data curation pipeline based on monocular videos.}\label{appendix: Details of data curation pipeline based on monocular videos}
We also implement an alternative pipeline based on monocular videos, as shown in Fig.~\ref{fig: monocular pipeline}. First, we employ Gemini-2.5-flash, to identify all moving entities and extract their corresponding noun categories. We then utilize GroundedSAM2 for instance segmentation and tracking, yielding temporally consistent 2D masks across the sequence. These semantic descriptions are further enriched using PixelRefer. We sample pixels in each region and apply a 4D tracking method, such as SpatialTrakerV2~\cite{xiao2025spatialtrackerv2} to track points in 4D space. Since 4D tracking method typically produces depth at a relative-scale, we incorporate a metric-scale depth estimation method, such as MoGe-2~\cite{wang2025moge}, to align the final per-frame object-level points. This monocular-based pipeline often faces challenges with depth ambiguity and diminished spatial accuracy inherent in 4D tracking and monocular depth estimation.

\subsection{Limitations and Future Work}
Although MLLM-4D demonstrates strong viusal-based spatial-temporal intelligence compared to existing MLLMs across various 4D reasoning tasks, it still faces constraints in processing long-duration video sequences. Due to the inherent input length limitations of current architectures, our model relies on frame sampling. An compelling direction for the future work would lies in exploring long-context 4D spatiotemporal reasoning.

\end{document}